\documentclass[10pt,twocolumn,letterpaper]{article}

\usepackage{cvpr}
\usepackage{times}
\usepackage{epsfig}
\usepackage{graphicx}
\usepackage{amsmath}
\usepackage{amssymb}
\usepackage{multicol}
\usepackage{enumitem}
\usepackage[breakable]{tcolorbox}

\setlist[itemize]{noitemsep,nolistsep}

\usepackage[breaklinks=true,bookmarks=false]{hyperref}

\cvprfinalcopy % *** Uncomment this line for the final submission

\def\cvprPaperID{****} % *** Enter the CVPR Paper ID here

\ifcvprfinal\fi
\begin{document}

%%%%%%%%% TITLE
\title{PULSE: Self-Supervised Photo Upsampling via \\Latent Space Exploration of Generative Models}

\author{Sachit Menon*, Alexandru Damian*, Shijia Hu, Nikhil Ravi, Cynthia Rudin\\
Duke University\\
Durham, NC\\
{\tt\small \{sachit.menon,alexandru.damian,shijia.hu,nikhil.ravi,cynthia.rudin\}@duke.edu}
}

\maketitle

\begin{abstract}
    % \vspace{-.228in}
   The primary aim of single-image super-resolution is to construct a high-resolution (HR) image from a corresponding low-resolution (LR) input. In previous approaches, which have generally been supervised, the training objective typically measures a pixel-wise average distance between the super-resolved (SR) and HR images. Optimizing such metrics often leads to blurring, especially in high variance (detailed) regions. We propose an alternative formulation of the super-resolution problem based on creating realistic SR images that downscale correctly. We present a novel super-resolution algorithm addressing this problem, PULSE (Photo Upsampling via Latent Space Exploration), which generates high-resolution, realistic images at resolutions previously unseen in the literature. It accomplishes this in an entirely self-supervised fashion and is not confined to a specific degradation operator used during training, unlike previous methods (which require training on databases of LR-HR image pairs for supervised learning). Instead of starting with the LR image and slowly adding detail, PULSE traverses the high-resolution natural image manifold, searching for images that downscale to the original LR image. This is formalized through the ``downscaling loss,'' which guides exploration through the latent space of a generative model. By leveraging properties of high-dimensional Gaussians, we restrict the search space to guarantee that our outputs are realistic. PULSE thereby generates super-resolved images that both are realistic and downscale correctly. We show extensive experimental results demonstrating the efficacy of our approach in the domain of face super-resolution (also known as face hallucination). We also present a discussion of the limitations and biases of the method as currently implemented with an accompanying model card with relevant metrics. Our method outperforms state-of-the-art methods in perceptual quality at higher resolutions and scale factors than previously possible. 
\end{abstract}

* denotes equal contribution

%%%%%%%%% BODY TEXT
\section{Introduction}
\begin{figure}[!ht]
    \centering
    \includegraphics[width=\columnwidth]{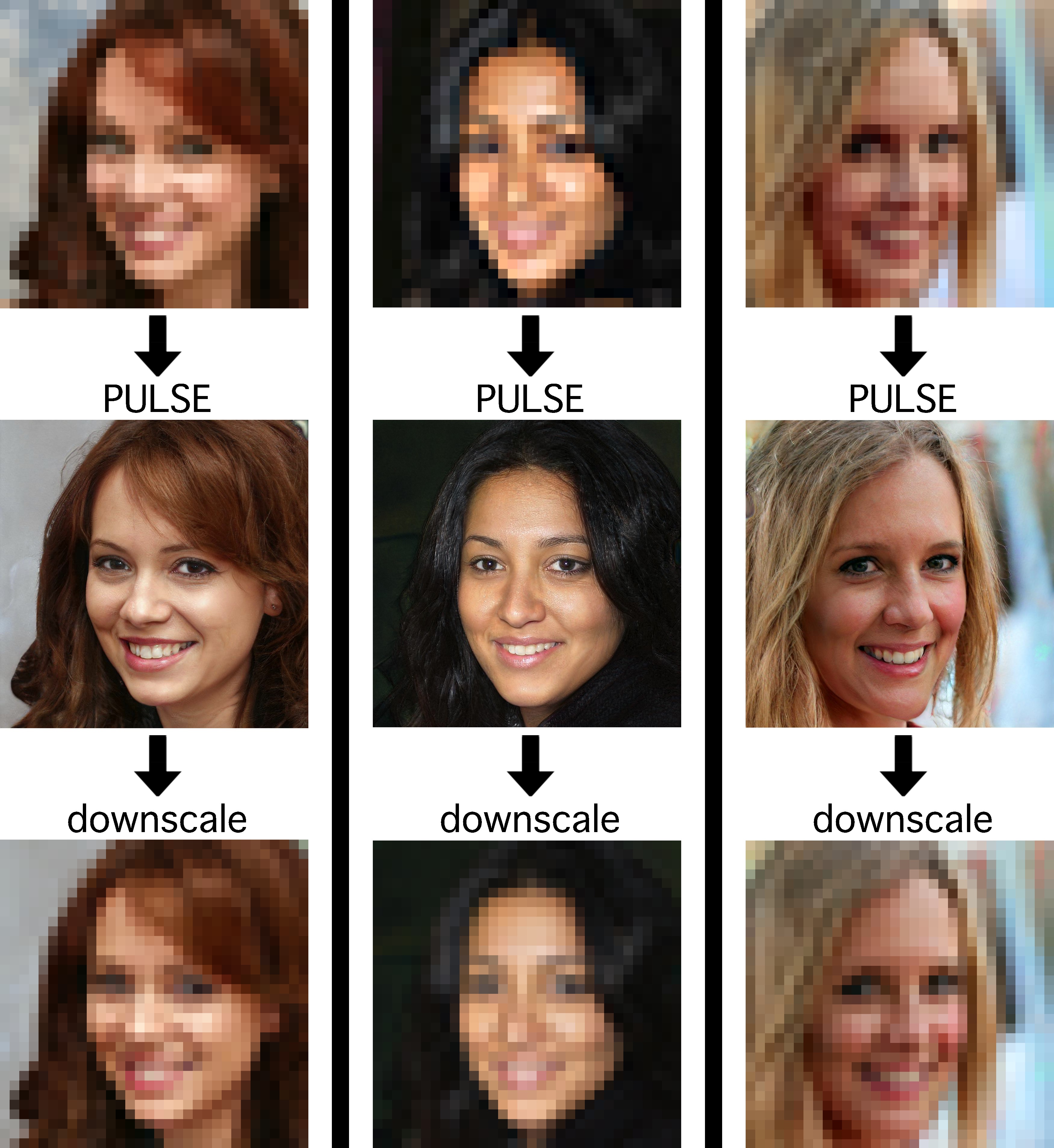}\\
    \caption{(x32) The input (top) gets upsampled to the SR image (middle) which downscales (bottom) to the original image.}
    \label{fig:process}
\end{figure}

\noindent In this work, we aim to transform blurry, low-resolution images into sharp, realistic, high-resolution images. Here, we focus on images of faces, but our technique is generally applicable. In many areas (such as medicine, astronomy, microscopy, and satellite imagery), sharp, high-resolution images are difficult to obtain due to issues of cost, hardware restriction, or memory limitations~\cite{singh2016super}. This leads to the capture of blurry, low-resolution images instead. In other cases, images could be old and therefore blurry, or even in a modern context, an image could be out of focus or a person could be in the background. In addition to being visually unappealing, this impairs the use of downstream analysis methods (such as image segmentation, action recognition, or disease diagnosis) which depend on having high-resolution images \cite{unet} \cite{actrec}. In addition, as consumer laptop, phone, and television screen resolution has increased over recent years, popular demand for sharp images and video has surged. This has motivated recent interest in the computer vision task of \textit{image super-resolution}, the creation of realistic high-resolution (henceforth HR) images that a given low-resolution (LR) input image could correspond to.

While the benefits of methods for image super-resolution are clear, the difference in information content between HR and LR images (especially at high scale factors) hampers efforts to develop such techniques. In particular, LR images inherently possess less high-variance information; details can be blurred to the point of being visually indistinguishable. The problem of recovering the true HR image depicted by an LR input, as opposed to generating a set of potential such HR images, is inherently ill-posed, as the size of the total set of these images grows exponentially with the scale factor~\cite{baker2000limits}. That is to say, \textit{many} high-resolution images can correspond to the exact same low-resolution image.

Traditional supervised super-resolution algorithms train a model (usually, a convolutional neural network, or CNN) to minimize the pixel-wise mean-squared error (MSE) between the generated super-resolved (SR) images and the corresponding ground-truth HR images~\cite{VDSR}~\cite{chen2018fsrnet}. However, this approach has been noted to neglect perceptually relevant details critical to photorealism in HR images, such as texture ~\cite{SRGAN}. Optimizing on an average difference in pixel-space between HR and SR images has a blurring effect, encouraging detailed areas of the SR image to be smoothed out to be, on average, more (pixelwise) correct. In fact, in the case of mean squared error (MSE), the ideal solution is the (weighted) pixel-wise average of the set of realistic images that downscale properly to the LR input (as detailed later). The inevitable result is smoothing in areas of high variance, such as areas of the image with intricate patterns or textures. As a result, MSE should not be used alone as a measure of image quality for super-resolution.

Some researchers have attempted to extend these MSE-based methods to additionally optimize on metrics intended to encourage realism, serving as a force opposing the smoothing pull of the MSE term~\cite{SRGAN,chen2018fsrnet}. This essentially drags the MSE-based solution in the direction of the natural image manifold (the subset of $\mathbb{R}^{M \times N}$ that represents the set of high-resolution images). This compromise, while improving perceptual quality over pure MSE-based solutions, makes no guarantee that the generated images are realistic. Images generated with these techniques still show signs of blurring in high variance areas of the images, just as in the pure MSE-based solutions.

\begin{figure}[!t]
    \centering
    \includegraphics[width=\columnwidth]{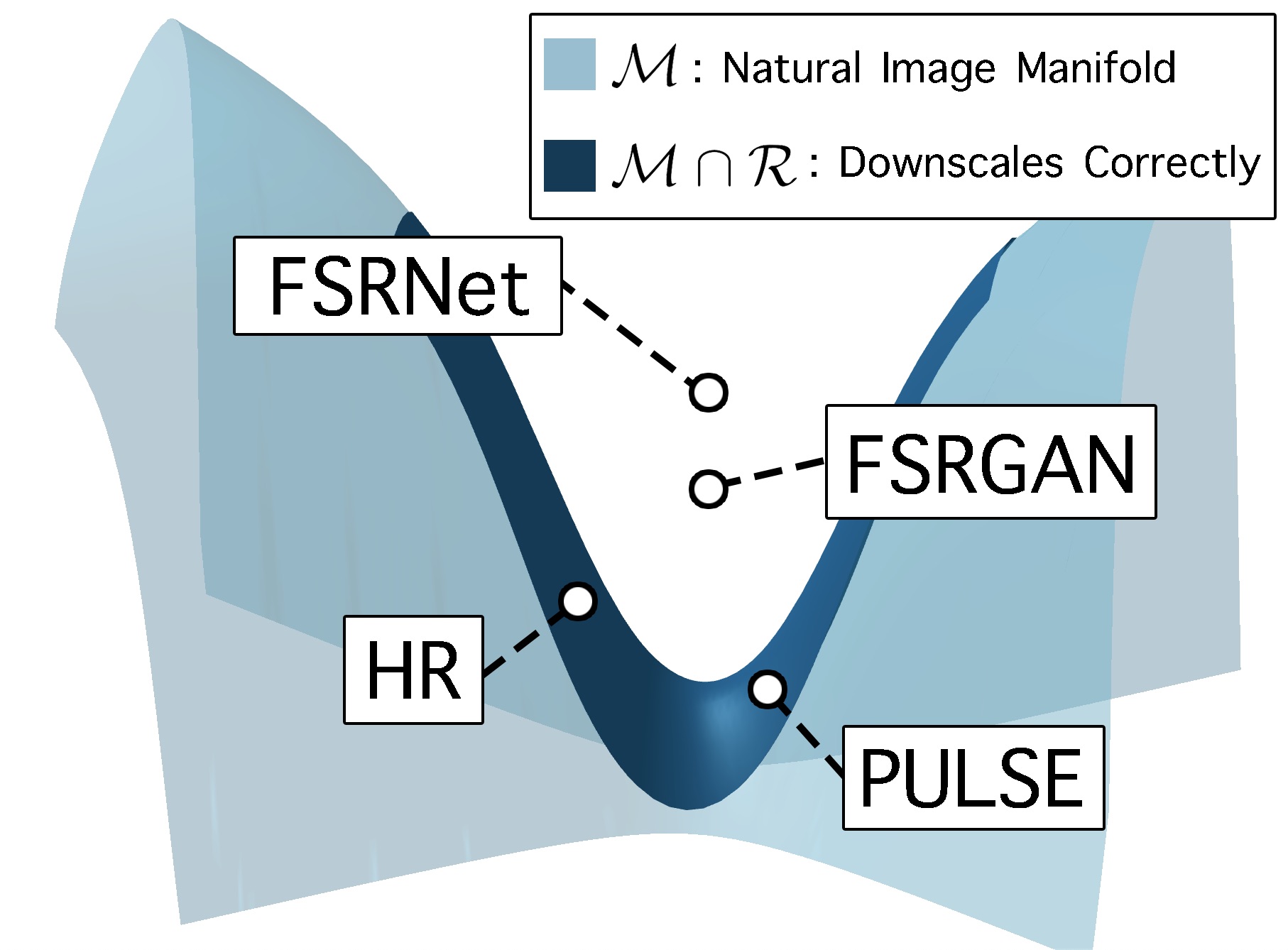}
    \caption{FSRNet tends towards an average of the images that downscale properly. The discriminator loss in FSRGAN pulls it in the direction of the natural image manifold, whereas PULSE always moves along this manifold.}
    \label{fig:manifold}
\end{figure}

To avoid these issues, we propose a new paradigm for super-resolution. The goal should be to generate realistic images within the set of feasible solutions; that is, to find points which \textit{actually lie on the natural image manifold and also downscale correctly}. The (weighted) pixel-wise average of possible solutions yielded by the MSE does not generally meet this goal for the reasons previously described. We provide an illustration of this in Figure \ref{fig:manifold}.

Our method generates images using a (pretrained) generative model approximating the distribution of natural images under consideration. For a given input LR image, we traverse the manifold, parameterized by the latent space of the generative model, to find regions that downscale correctly. In doing so, we find examples of realistic images that downscale properly, as shown in \ref{fig:process}.

Such an approach also eschews the need for supervised training, being entirely self-supervised with no `training' needed at the time of super-resolution inference (except for the unsupervised generative model). This framework presents multiple substantial benefits. First, it allows the same network to be used on images with differing degradation operators even in the absence of a database of corresponding LR-HR pairs (as no training on such databases takes place). Furthermore, unlike previous methods, it does not require super-resolution task-specific network architectures, which take substantial time on the part of the researcher to develop without providing real insight into the problem; instead, it proceeds alongside the state-of-the-art in generative modeling, with zero retraining needed.

Our approach works with any type of generative model with a differentiable generator, including flow-based models, variational autoencoders (VAEs), and generative adversarial networks (GANs); the particular choice is dictated by the tradeoffs each make in approximating the data manifold. For this work, we elected to use GANs due to recent advances yielding high-resolution, sharp images \cite{karras2019style,karras2017progressive}. 

One particular subdomain of image super-resolution deals with the case of face images. This subdomain -- known as \textit{face hallucination} -- finds application in consumer photography, photo/video restoration, and more \cite{srsurvey}. As such, it has attracted interest as a computer vision task in its own right. Our work focuses on face hallucination, but our methods extend to a more general context.

Because our method always yields a solution that both lies on the natural image manifold and downsamples correctly to the original low-resolution image, we can provide a range of interesting high-resolution possibilities e.g. by making use of the stochasticity inherent in many generative models: our technique can create a \textit{set} of images, \textit{each} of which is visually convincing, yet look different from each other, where (without ground truth) \textit{any} of the images could plausibly have been the source of the low-resolution input.

\begin{figure}[!t]
    \centering
    \includegraphics[width=0.8\columnwidth ]{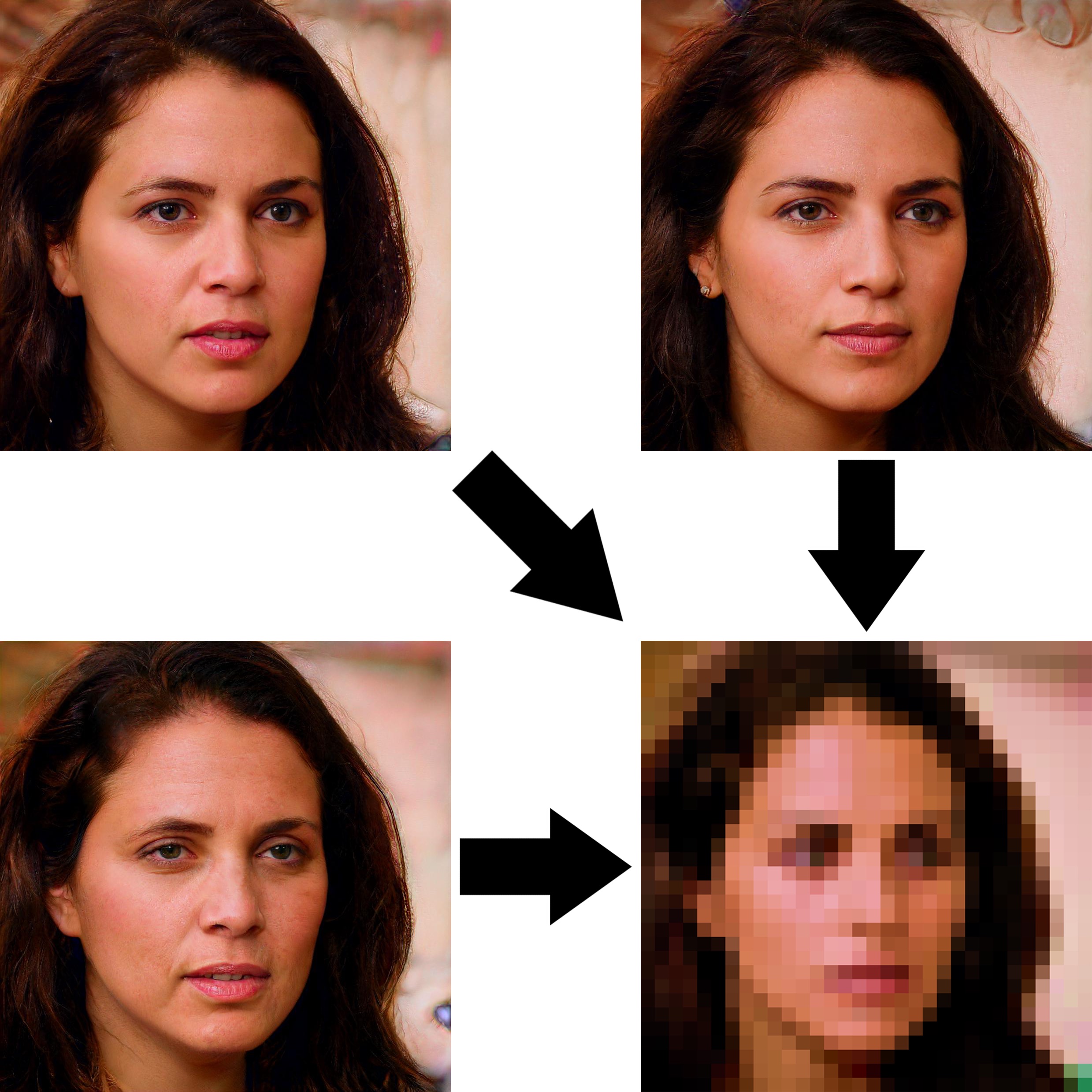}
    \caption{We show here how visually distinct images, created with PULSE, can all downscale (represented by the arrows) to the same LR image. }
    \label{fig:multipledownscale}
\end{figure}

Our main contributions are as follows.
\begin{enumerate}
    \item \textbf{A new paradigm for image super-resolution}. Previous efforts take the traditional, ill-posed perspective of attempting to `reconstruct' an HR image from an LR input, yielding outputs that, in effect, average many possible solutions. This averaging introduces undesirable blurring. We introduce new approach to super-resolution: a super-resolution algorithm should create realistic high-resolution outputs that downscale to the correct LR input.
    \item \textbf{A novel method for solving the super-resolution task}. In line with our new perspective, we propose a new algorithm for super-resolution. Whereas traditional work has at its core aimed to approximate the LR $\rightarrow$ HR map using supervised learning (especially with neural networks), our approach centers on the use of unsupervised generative models of HR data. Using generative adversarial networks, we explore the latent space to find regions that map to realistic images and downscale correctly. No retraining is required. Our particular implementation, using StyleGAN \cite{karras2019style}, allows for the creation of any number of realistic SR samples that correctly map to the LR input.
    \item \textbf{An original method for latent space search under high-dimensional Gaussian priors.} In our task and many others, it is often desirable to find points in a generative model's latent space that map to realistic outputs. Intuitively, these should resemble samples seen during training. At first, it may seem that traditional log-likelihood regularization by the latent prior would accomplish this, but we observe that the `soap bubble' effect (that much of the density of a high dimensional Gaussian lies close to the surface of a hypersphere) contradicts this. Traditional log-likelihood regularization actually tends to draw latent vectors away from this hypersphere and, instead, towards the origin. We therefore constrain the search space to the surface of that hypersphere, which ensures realistic outputs in higher-dimensional latent spaces; such spaces are otherwise difficult to search.

    % However, we show these do not correspond to points with high prior likelihood in the latent space; instead, observing the `soap bubble' effect of Gaussian densities in high dimensions, we constrain the search space to the surface of a hypersphere where density collects. This ensures realistic outputs in higher-dimensional latent spaces, which are otherwise infeasible to search. 
    % \item \textbf{The introduction of a new loss term core to the super-resolution problem}. Current metrics for assessing the quality of super-resolution algorithms roughly fall into two categories: those evaluating correctness and those estimating perceptual quality. However, as stated in point 1 above, thus far `correctness' has been interpreted as `average pixel distance from original HR,' leading to metrics such as peak signal-to-noise ratio (PSNR). Many have noted these objectives run counter to the true goals of image super-resolution. A potentially more useful perspective is to consider correctness as how well the generated output downscales to the appropriate LR image. We propose the \textit{downscaling loss} for this purpose. 
\end{enumerate}

\section{Related Work}

While there is much work on image super-resolution prior to the advent of convolutional neural networks (CNNs), CNN-based approaches have rapidly become state-of-the-art in the area and are closely relevant to our work; we therefore focus on neural network-based approaches here. Generally, these methods use a pipeline where a low-resolution (LR) image, created by down-sampling a high-resolution (HR) image, is fed through a CNN with both convolutional and upsampling layers, generating a super-resolved (SR) output. This output is then used to calculate the loss using the chosen loss function and the original HR image.

\subsection{Current Trends}
Recently, supervised neural networks have come to dominate current work in super-resolution. Dong \textit{et al.} \cite{10.1007/978-3-319-10593-2_13} proposed the first CNN architecture to learn this non-linear LR to HR mapping using pairs of HR-LR images. Several groups have attempted to improve the upsampling step by utilizing sub-pixel convolutions and transposed convolutions~\cite{Shi_2016_CVPR}. Furthermore, the application of ResNet architectures to super-resolution (started by SRResNet \cite{SRGAN}), has yielded substantial improvement over more traditional convolutional neural network architectures. In particular, the use of residual structures allowed for the training of larger networks. Currently, there exist two general trends: one, towards networks that primarily better optimize pixel-wise average distance between SR and HR, and two, networks that focus on perceptual quality.

\subsection{Loss Functions}
Towards these different goals, researchers have designed different loss functions for optimization that yield images closer to the desired objective. Traditionally, the loss function for the image super-resolution task has operated on a per-pixel basis, usually using the L2 norm of the difference between the ground truth and the reconstructed image, as this directly optimizes PSNR (the traditional metric for the super-resolution task). More recently, some researchers have started to use the L1 norm since models trained using L1 loss seem to perform better in PSNR evaluation. The L2 norm (as well as pixel-wise average distances in general) between SR and HR images has been heavily criticized for not correlating well with human-observed image quality~\cite{SRGAN}. In face super-resolution, the state-of-the-art for such metrics is FSRNet~\cite{chen2018fsrnet}, which used a facial prior to achieve previously unseen PSNR.

Perceptual quality, however, does not necessarily increase with higher PSNR. As such, different methods, and in particular, objective functions, have been developed to increase perceptual quality. In particular, methods that yield high PSNR result in blurring of details. The information required for details is often not present in the LR image and must be `imagined' in. One approach to avoiding the direct use of the standard loss functions was demonstrated in \cite{deep_prior}, which draws a prior from the structure of a convolutional network. This method produces similar images to the methods that focus on PSNR, which lack detail, especially in high frequency areas. Because this method cannot leverage learned information about what realistic images look like, it is unable to fill in missing details. Methods that try to learn a map from LR to HR images can try to leverage learned information; however, as mentioned, networks optimized on PSNR are still explicitly penalized for attempting to hallucinate details they are unsure about, thus optimizing on PSNR stills resulting in blurring and lack of detail. 
% This is a distinct problem from the one that traditional methods that learn a map from LR to HR images and optimize on PSNR face. These methods can leverage the training data to learn what realistic images look like, but are penalized by the loss term for attempting to hallucinate details they are unsure about.

To resolve this issue, some have tried to use generative model-based loss terms to provide these details. Neural networks have lent themselves to application in generative models of various types (especially generative adversarial networks--GANs--from \cite{goodfellow2014generative}), to image reconstruction tasks in general, and more recently, to super-resolution. Ledig \textit{et al.} \cite{SRGAN} created the SRGAN architecture for single-image upsampling by leveraging these advances in deep generative models, specifically GANs. Their general methodology was to use the generator to upscale the low-resolution input image, which the discriminator then attempts to distinguish from real HR images, then propagate the loss back to both networks. Essentially, this optimizes a supervised network much like MSE-based methods with an additional loss term corresponding to how fake the discriminator believes the generated images to be. However, this approach is fundamentally limited as it essentially results in an averaging of the MSE-based solution and a GAN-based solution, as we discuss later. In the context of faces, this technique has been incorporated into FSRGAN, resulting in the current perceptual state-of-the-art in face super resolution at $\times 8$ upscaling factors up to resolutions of $128 \times 128$. Although these methods use a `generator' and a `discriminator' as found in GANs, they are trained in a completely supervised fashion; they do not use unsupervised generative models. 
%No methods in SR thus far utilize a truly unsupervised generative model to our knowledge. 

% Other novel loss functions have emerged besides the perceptual loss function used to train SRGAN that depend on the network architecture. The LapSRN~\cite{LapSRN} algorithm introduced a pyramid network architecture where each layer progressively predicts high-frequency residuals from coarse feature maps. The loss functions for each layer of the pyramid uses the L1 norm and the residual image at a particular layer to model the desired HR output.

\subsection{Generative Networks}

% \textcolor{red}{suggested rewording: Our algorithm takes a GAN (or other generative model) and searches its latent space for latents that map to images that downscale correctly.}
Our algorithm does not simply use GAN-style training; rather, it uses a truly unsupervised GAN (or, generative model more broadly). It searches the latent space of this generative model for latents that map to images that downscale correctly. The quality of cutting-edge generative models is therefore of interest to us. 
% , and searches its latent space for latents that map to images that downscale correctly. 

As GANs have produced the highest-quality high-resolution images of deep generative models to date, we chose to focus on these for our implementation. Here we provide a brief review of relevant GAN methods with high-resolution outputs. Karras \textit{et al}. \cite{karras2017progressive} presented some of the first high-resolution outputs of deep generative models in their ProGAN algorithm, which grows both the generator and the discriminator in a progressive fashion. Karras \textit{et al}. \cite{karras2019style} further built upon this idea with StyleGAN, aiming to allow for more control in the image synthesis process relative to the black-box methods that came before it. The input latent code is embedded into an intermediate latent space, which then controls the behavior of the synthesis network with adaptive instance normalization applied at each convolutional layer. This network has 18 layers (2 each for each resolution from 4 $\times$ 4 to 1024 $\times$ 1024). After every other layer, the resolution is progressively increased by a factor of 2. At each layer, new details are introduced stochastically via Gaussian input to the adaptive instance normalization layers. Without perturbing the discriminator or loss functions, this architecture leads to the option for scale-specific mixing and control over the expression of various high-level attributes and variations in the image (e.g. pose, hair, freckles, etc.). Thus, StyleGAN provides a very rich latent space for expressing different features, especially in relation to faces. 

% \subsection{Face Super-resolution (aka Face Hallucination)}
% Face hallucination, super-resolution specifically on faces, has seen much progress recently. One of the most profound breakthroughs was FSRNet, created by Chen et al. (CITATION). One of the main improvements with their technique was to include a facial segmentation map and a separate loss term, which they term a facial prior, consisting of the difference between facial landmark heatmaps. This allowed them to achieve state of the art performance in face hallucination. However, they only accomplish this with a highest resolution of 128x128 pixels. 

% Yu and Porikli provide a different approach with their URDGN model [CITATION]. This architecture first uses a transformative discriminative decoder to upsamle and denoise the LR image, a transformative encoder to project the output onto the LR image to align and remove artifacts, and finally another decoder to generate the final, super-resolved images (follows the decoder-encoder-decoder framework). 

% Kim et al. chose to employ a progressive method in training their face hallucination model. Each successive step in their network produces an output with progressively higher resolutions. They also created a "facial attention loss" which focuses on preserving feature detail in important facial landmarks, which is used at every step of the network.

\section{Method}
We begin by defining some universal terminology necessary to any formal description of the super-resolution problem. We denote the low-resolution input image by $I_{LR}$. 
We aim to learn a conditional generating function $G$ that, when applied to $I_{LR}$, yields a higher-resolution super-resolved image $I_{SR}$. Formally, let $I_{LR} \in \mathbb{R}^{m \times n}$. Then our desired function $SR$ is a map $\mathbb{R}^{m \times n} \rightarrow \mathbb{R}^{M \times N}$ where $M > m$, $N > n$. We define the super-resolved image $I_{SR} \in \mathbb{R}^{M \times N}$
\begin{align}
    I_{SR} &:= SR(I_{LR}).
\end{align}

In a traditional approach to super-resolution, one considers that the low-resolution image could represent the same information as a theoretical high-resolution image $I_{HR} \in \mathbb{R}^{M \times N}$. The goal is then to best recover this particular $I_{HR}$ given $I_{LR}$. Such approaches therefore reduce the problem to an optimization task: fit a function $SR$ that minimizes
\begin{align}
    L:= \|I_{HR} - I_{SR}\|_p^p
\end{align}
where $\| \cdot \|_p$ denotes some $l^p$ norm. 

In practice, even when trained correctly, these algorithms fail to enhance detail in high variance areas. To see why this is, fix a low resolution image $I_{LR}$. Let $\mathcal{M}$ be the natural image manifold in $\mathbb{R}^{M \times N}$, i.e., the subset of $\mathbb{R}^{M \times N}$ that resembles natural realistic images, and let $P$ be a probability distribution over $\mathcal{M}$ describing the likelihood of an image appearing in our dataset. Finally, let $R$ be the set of images that downscale correctly, i.e., $R = \{I \in \mathbb{R}^{N\times M} ~:~ DS(I)=I_{LR}\}$. Then in the limit as the size of our dataset tends to infinity, our expected loss when the algorithm outputs a fixed image $I_{SR}$ is
\begin{align}
\int_{\mathcal{M} \cap R} \|I_{HR} - I_{SR}\|_p^p ~ dP(I_{HR}).
\end{align}
This is minimized when $I_{SR}$ is an $l_p$ average of $I_{HR}$ over $M \cap R$. In fact, when $p=2$, this is minimized when
\begin{align}
I_{SR} = \int_{\mathcal{M} \cap R} I_{HR} ~ dP(I_{HR}),
\end{align}
so \textit{the optimal $I_{SR}$ is a weighted pixelwise average of the set of high resolution images that downscale properly}. As a result, the lack of detail in algorithms that rely only on an $l_p$ norm cannot be fixed simply by changing the architecture of the network. The problem itself has to be rephrased.
\begin{figure}[!t]
    \centering
    \includegraphics[width=0.95\columnwidth]{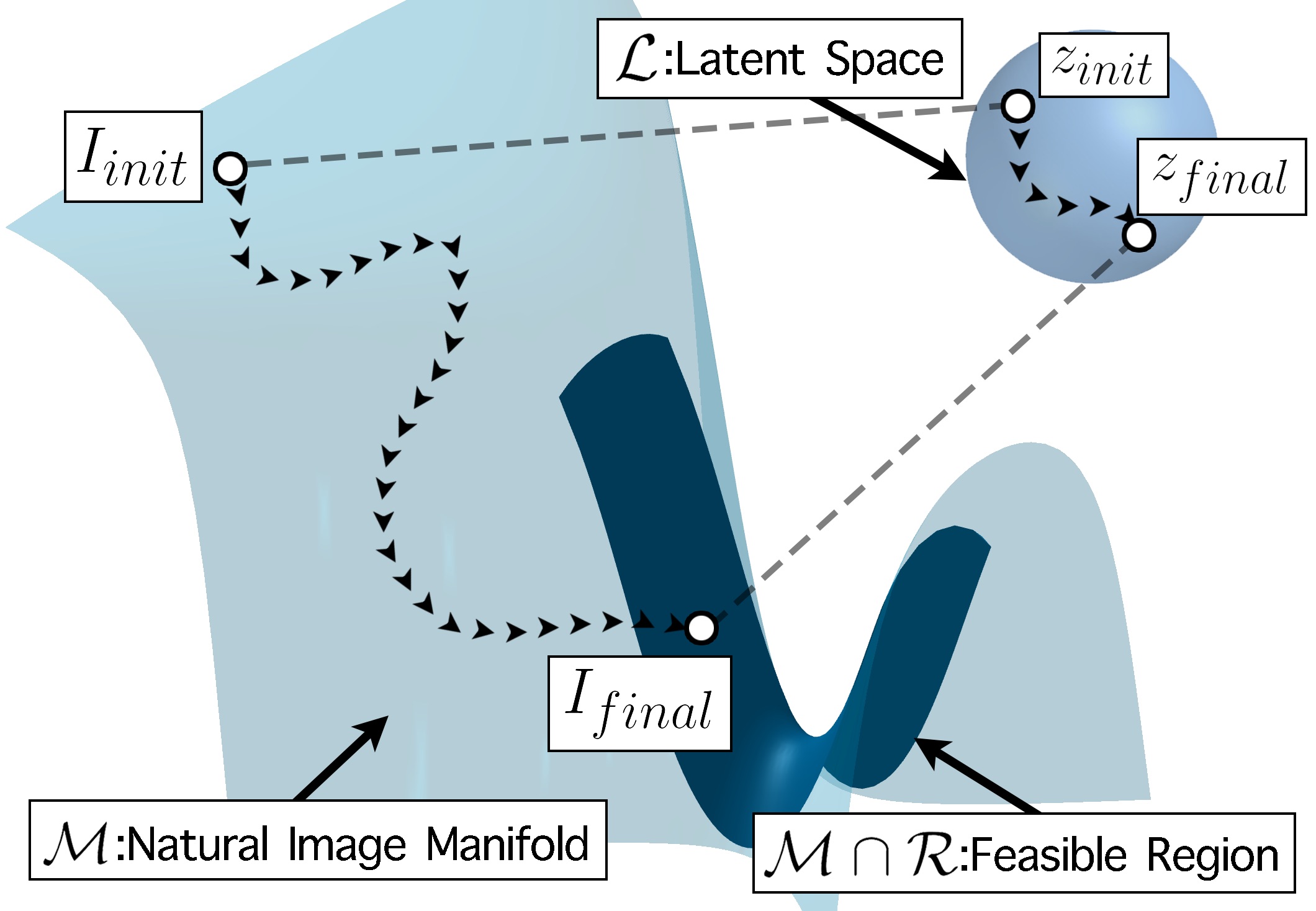}
    \caption{While traveling from $z_{init}$ to $z_{final}$ in the latent space $\mathcal{L}$, we travel from $I_{init} \in \mathcal{M}$ to $I_{final} \in \mathcal{M} \cap \mathcal{R}$.}
    \label{fig:algorithm}
\end{figure}

We therefore propose a new framework for single image super resolution. Let $\mathcal{M}$, $DS$ be defined as above. Then for a given LR image $I_{LR} \in \mathbb{R}^{m \times n}$ and $\epsilon > 0$, our goal is to find an image $I_{SR} \in \mathcal{M}$ with
\begin{align}
\|DS(I_{SR})-I_{LR}\|_p \leq \epsilon.
\end{align} In particular, we can let $\mathcal{R}_\epsilon \subset \mathbb{R}^{N\times M}$ be the set of images that downscale properly, i.e.,
\begin{align}
\mathcal{R}_\epsilon = \{I \in \mathbb{R}^{N\times M} ~:~ \|DS(I)-I_{LR}\|_p^p \leq \epsilon\}.
\end{align} Then we are seeking an image $I_{SR} \in \mathcal{M} \cap \mathcal{R}_\epsilon$. The set $\mathcal{M} \cap \mathcal{R}_\epsilon$ is the set of \textit{feasible solutions}, because a solution is not feasible if it did not downscale properly and look realistic.

It is also interesting to note that the intersections $\mathcal{M} \cap \mathcal{R}_\epsilon$ and in particular $\mathcal{M} \cap \mathcal{R}_0$ are guaranteed to be nonempty, because they must contain the original HR image (i.e., what traditional methods aim to reconstruct).

\subsection{Downscaling Loss}
Central to the problem of super-resolution, unlike general image generation, is the notion of \textit{correctness}. Traditionally, this has been interpreted to mean how well a particular ground truth image $I_{HR}$ is `recovered' by the application of the super-resolution algorithm $SR$ to the low-resolution input $I_{LR}$, as discussed in the related work section above. This is generally measured by some $l_p$ norm between $I_{SR}$ and the ground truth, $I_{HR}$; such algorithms only look somewhat like real images because minimizing this metric drives the solution somewhat nearer to the manifold. However, they have no way to ensure that $I_{SR}$ lies \textit{close} to $\mathcal{M}$. In contrast, in our framework, we never deviate from $\mathcal{M}$, so such a metric is not necessary. For us, the critical notion of correctness is how well the generated SR image $I_{SR}$ corresponds to $I_{LR}$.

We formalize this through the \textit{downscaling loss}, to explicitly penalize a proposed SR image for deviating from its LR input (similar loss terms have been proposed in \cite{styleganembedding},\cite{deep_prior}). This is inspired by the following: for a proposed SR image to represent the same information as a given LR image, it must downscale to this LR image. That is,
\begin{align}
    I_{LR} \approx DS(I_{SR}) = DS(SR(I_{LR}))
\end{align}
where $DS(\cdot)$ represents the downscaling function.

Our downscaling loss therefore penalizes $SR$ the more its outputs violate this,
\begin{align}
    L_{DS}(I_{SR},I_{LR}):= \| DS(I_{SR}) - I_{LR}\|_p^p.
\end{align}

It is important to note that the downscaling loss can be used in both supervised and unsupervised models for super-resolution; it does not depend on an HR reference image.

\subsection{Latent Space Exploration} \label{lse}
How might we find regions of the natural image manifold $\mathcal{M}$ that map to the correct LR image under the downscaling operator? If we had a differentiable parameterization of the manifold, we could progress along the manifold to these regions by using the downscaling loss to guide our search. In that case, images found would be guaranteed to be high resolution as they came from the HR image manifold, while also being correct as they would downscale to the LR input. 

In reality, we do not have such convenient, perfect parameterizations of manifolds. However, we can approximate such a parameterization by using techniques from unsupervised learning. In particular, much of the field of deep generative modeling (e.g. VAEs, flow-based models, and GANs) is concerned with creating models that map from some latent space to a given manifold of interest. By leveraging advances in generative modeling, we can even use pretrained models without the need to train our own network. Some prior work has aimed to find vectors in the latent space of a generative model to accomplish a task; see \cite{styleganembedding} for creating embeddings and \cite{bora2017compressed} in the context of compressed sensing. (However, as we describe later, this work does not actually search in a way that yields realistic outputs as intended.) In this work, we focus on GANs, as recent work in this area has resulted in the highest quality image-generation among unsupervised models.

Regardless of its architecture, let the generator be called $G$, and let the latent space be $\mathcal{L}$. Ideally,  we could approximate $\mathcal{M}$ by the image of $G$, which would allow us to rephrase the problem above as the following: find a latent vector $z \in \mathcal{L}$ with 
\begin{align}
\label{eq:dsloss}
    \|DS(G(z))-I_{LR}\|_p^p \leq \epsilon.
\end{align} 
Unfortunately, in most generative models, simply requiring that $z \in \mathcal{L}$ does not guarantee that $G(z) \in \mathcal{M}$; rather, such methods use an imposed prior on $\mathcal{L}$. In order to ensure $G(z) \in \mathcal{M}$, we must be in a region of $\mathcal{L}$ with high probability under the chosen prior. One idea to encourage the latent to be in the region of high probability is to add a loss term for the negative log-likelihood of the prior. In the case of a Gaussian prior, this takes the form of $l_2$ regularization. Indeed, this is how the previously mentioned work \cite{bora2017compressed} attempts to address this issue. However, this idea does not actually accomplish the goal. Such a penalty forces vectors towards $0$, but most of the mass of a high-dimensional Gaussian is located near the surface of a sphere of radius $\sqrt{d}$ (see \cite{vershynin_2018}). To get around this, we observed that we could replace the Gaussian prior on $\mathbb{R}^{d}$ with a uniform prior on $\sqrt{d}\mathbb{S}^{d-1}$. This approximation can be used for any method with high dimensional spherical Gaussian priors.
%The aforementioned StyleGAN and ProGAN both make use of this, sampling from the surface of a sphere in lieu of a Gaussian. 

We can let $\mathcal{L}'=\sqrt{d}S^{d-1}$ (where $S^{d-1} \subset \mathbb{R}^{d}$ is the unit sphere in $d$ dimensional Euclidean space) and reduce the problem above to finding a $z \in \mathcal{L}'$ that satisfies Equation (\ref{eq:dsloss}). This reduces the problem from gradient descent in the entire latent space to projected gradient descent on a sphere.

\section{Experiments}

We designed various experiments to assess our method. We focus on the popular problem of face hallucination, enhanced by recent advances in GANs applied to face generation. In particular, we use Karras \textit{et al.}'s pretrained Face StyleGAN (trained on the Flickr Face HQ Dataset, or FFHQ) \cite{karras2019style}. We adapted the implementation found at \cite{PyTorchStyleGAN} in order to transfer the original StyleGAN-FFHQ weights and model from TensorFlow \cite{abadi2016tensorflow} to PyTorch \cite{paszke2019pytorch}. For each experiment, we used $100$ steps of spherical gradient descent with a learning rate of $0.4$ starting with a random initialization. Each image was therefore generated in ${\sim}5$ seconds on a single NVIDIA V100 GPU. 

\subsection{Data}
We evaluated our procedure on the well-known high-resolution face dataset CelebA HQ. (Note: this is not to be confused with CelebA, which is of substantially lower resolution.) We performed these experiments using scale factors of $64 \times$, $32\times$, and $8\times$.
% (that is, a $1024\times$, $256\times$, and $64\times$ reduction in pixels respectively)
For our qualitative comparisons, we upscale at scale factors of both $8 \times$ and $64 \times$, i.e., from $16 \times 16$ to $128 \times 128$ resolution images and $1024 \times 1024$ resolution images. The state-of-the-art for face super-resolution in the literature prior to this point was limited to a maximum of $8 \times$ upscaling to a resolution of $128 \times 128$, thus making it impossible to directly make quantitative comparisons at high resolutions and scale factors. We followed the traditional approach of training the supervised methods on CelebA HQ. We tried comparing with supervised methods trained on FFHQ, but they failed to generalize and yielded very blurry and distorted results when evaluated on CelebA HQ; therefore, in order to compare our method with the best existing methods, we elected to train the supervised models on CelebA HQ instead of FFHQ. 

%As techniques aimed at enhancing perceptual quality are inherently difficult to quantify, we begin with qualitative comparisons. \textcolor{red}{INCLUDE EVERYTHING WE END UP PROVIDING}.
\begin{figure*}[!ht]
    \centering
    \includegraphics[width=0.9\textwidth]{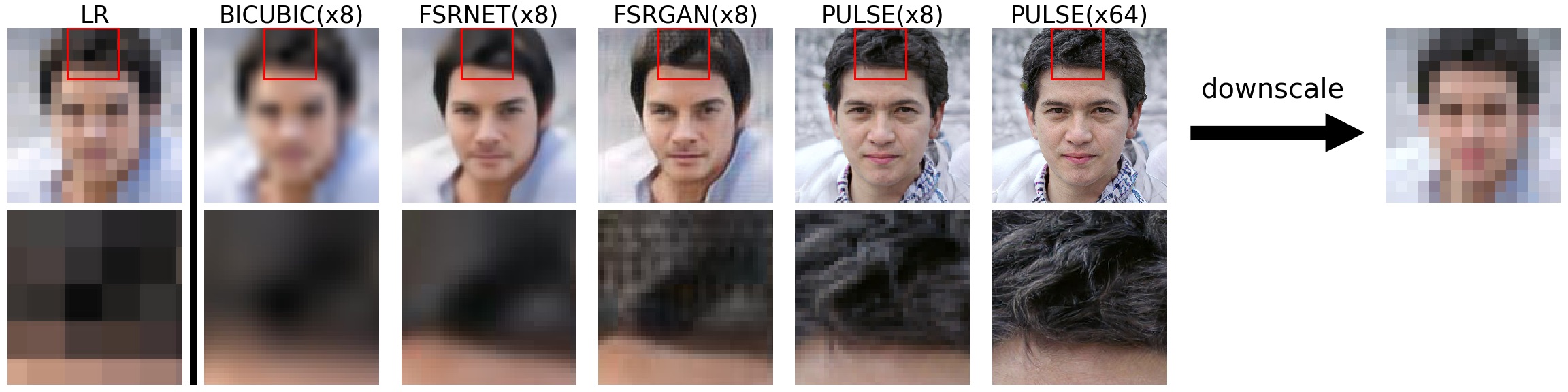}
    \includegraphics[width=0.9\textwidth]{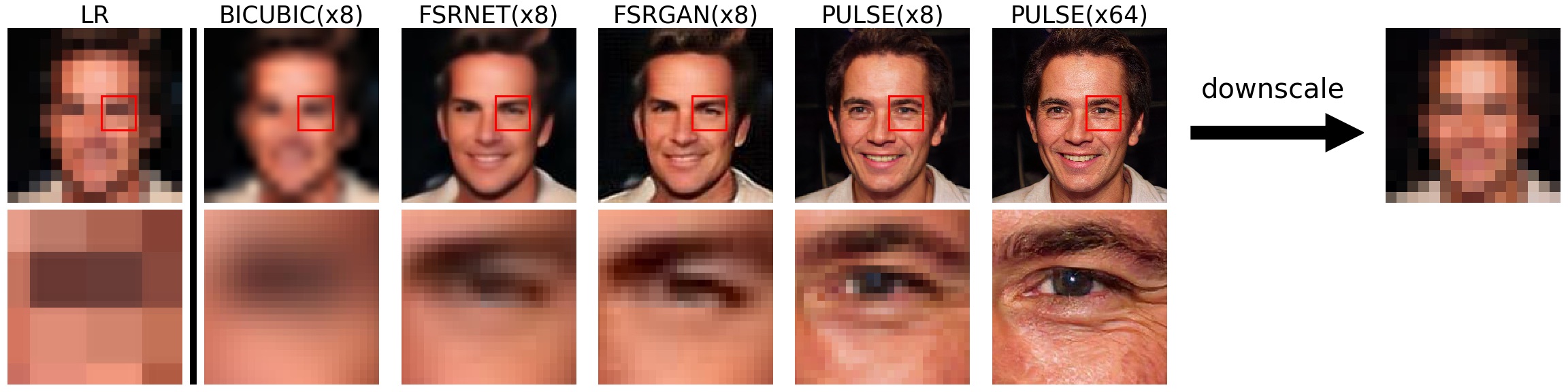}
    \includegraphics[width=0.9\textwidth]{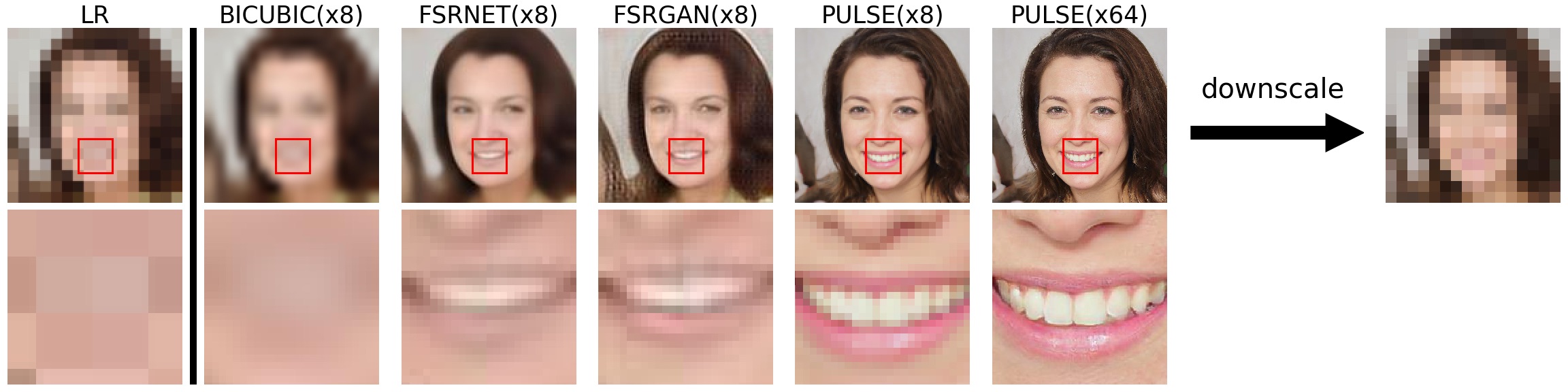}
    \caption{Comparison of PULSE with bicubic upscaling, FSRNet, and FSRGAN. In the first image, PULSE adds a messy patch in the hair to match the two dark diagonal pixels visible in the middle of the zoomed in LR image.}
    \label{fig:algcomparison}
\end{figure*}

\subsection{Qualitative Image Results}
Figure \ref{fig:algcomparison} shows qualitative results to demonstrate the visual quality of the images from our method. We observe levels of detail that far surpass competing methods, as exemplified by certain high frequency regions (features like eyes or lips). More examples and full-resolution images are in the appendix.

\subsection{Quantitative Comparison} \label{quantcomp}
Here we present a quantitative comparison with state-of-the-art face super-resolution methods. Due to constraints on the peak resolution that previous methods can handle, evaluation methods were limited, as detailed below. 
%Again, this is based on x8 upscaling to a resolution of $128 \times 128$ ($64 \times$ less pixels). 

We conducted a mean-opinion-score (MOS) test as is common in the perceptual super-resolution literature \cite{SRGAN,kim2019progressive}. For this, we had 40 raters examine images upscaled by 6 different methods (nearest-neighbors, bicubic, FSRNet, FSRGAN, and our PULSE). For this comparison, we used a scale factor of $8$ and a maximum resolution of $128 \times 128$, despite our method's ability to go substantially higher, due to this being the maximum limit for the competing methods. After being exposed to 20 examples of a 1 (worst) rating exemplified by nearest-neighbors upsampling, and a 5 (best) rating exemplified by high-quality HR images, raters provided a score from 1-5 for each of the 240 images. All images fell within the appropriate $\epsilon = 1e-3$ for the downscaling loss. The results are displayed in Table \ref{tab:MOS}. 
\begin{table}[]
\small{
\renewcommand{\arraystretch}{1.2}
\begin{tabular}{|c|c|c|c|c|c|}
\hline
HR   & Nearest & Bicubic & FSRNet & FSRGAN & PULSE \\ \hline \hline
3.74 & 1.01    & 1.34    & 2.77   & 2.92   & \textbf{3.60}  \\ \hline
\end{tabular}
}
\centering
\caption{MOS Score for various algorithms at $128 \times 128$. Higher is better.}
\label{tab:MOS}
\end{table}

PULSE outperformed the other methods and its score approached that of the HR dataset. Note that the HR's 3.74 average image quality reflects the fact that some of the HR images in the dataset had noticeable artifacts. All pairwise differences were highly statistically significant ($p < 10^{-5}$ for all 15 comparisons) by the Mann-Whitney-U test. The results demonstrate that PULSE outperforms current methods in generating perceptually convincing images that downscale correctly. 

To provide another measure of perceptual quality, we evaluated the Naturalness Image Quality Evaluator (NIQE) score \cite{NIQE}, previously used in perceptual super-resolution \cite{jeong2015multi,Blau_2018_ECCV_Workshops,esrgan}. This no-reference metric extracts features from images and uses them to compute a perceptual index (lower is better). As such, however, it only yields meaningful results at higher resolutions. This precluded direct comparison with FSRNet and FSRGAN, which produce images of at most $128 \times 128$ pixels.
% To provide some notion of comparison, we applied traditional bicubic upsampling to the output of these networks to $1024 \times 1024$ so as to include them here, but we note that this is not meant to be a direct comparison (as again, these previous networks are not capable of such). 

We evaluated NIQE scores for each method at a resolution of $1024 \times 1024$ from an input resolution of $16 \times 16$, for a scale factor of $64$. All images for each method fell within the appropriate $\epsilon = 1e-3$ for the downscaling loss. The results are in Table \ref{tab:NIQE}.
\begin{table}[]
\small{
\renewcommand{\arraystretch}{1.2}
\begin{tabular}{|c|c|c|c|c|c|}
\hline
HR   & Nearest & Bicubic & PULSE \\ \hline \hline
3.90 & 12.48 & 7.06 & \textbf{2.47}  \\ \hline
\end{tabular}
}
\centering
\caption{NIQE Score for various algorithms at $1024 \times 1024$. Lower is better.}
\label{tab:NIQE}
\end{table}
PULSE surpasses even the CelebA HQ images in terms of NIQE here, further showing the perceptual quality of PULSE's generated images. This is possible as NIQE is a no-reference metric which solely considers perceptual quality; unlike reference metrics like PSNR, performance is not bounded above by that of the HR images typically used as reference.

% \subsubsection{Downscaling Loss}

% \textcolor{red}{[INSERT DOWNSCALING LOSS TABLE HERE]}

% While the downscaling losses of the methods compared are comparable, each falls within the acceptable $\epsilon$-correctness range. 

\subsection{Image Sampling}

As referenced earlier, we initialize the point we start at in the latent space by picking a random point on the sphere. We found that we did not encounter any issues with convergence from random initializations. In fact, this provided us one method of creating many different outputs with high-level feature differences: starting with different initializations. An example of the variation in outputs yielded by this process can be observed in Figure \ref{fig:multipledownscale}.

Furthermore, by utilizing a generative model with inherent stochasticity, we found we could sample faces with fine-level variation that downscale correctly; this procedure can be repeated indefinitely. In our implementation, we accomplish this by resampling the noise inputs that StyleGAN uses to fill in details within the image.
% \subsection{Noisy Images}

% Our method also performs successfully in the presence of additive noise. 

\begin{figure}[]
    \centering
    \includegraphics[width=1\columnwidth]{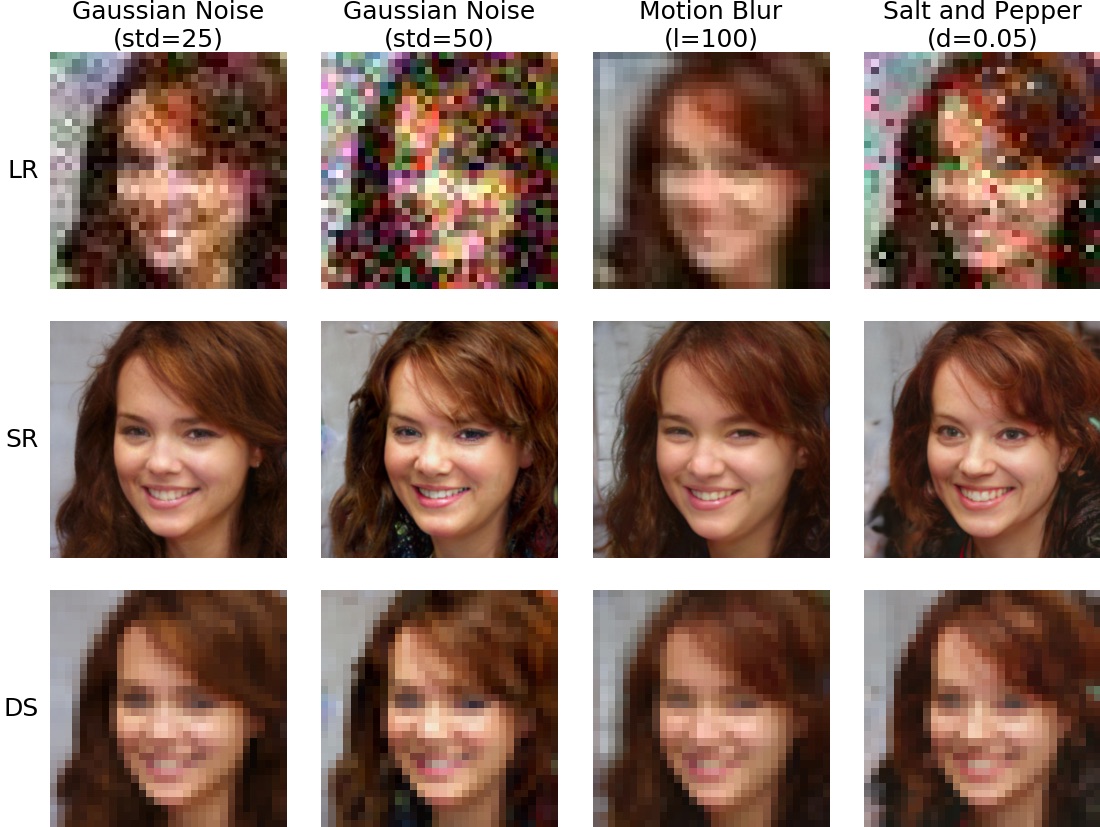}
    \caption{(x32) We show the robustness of PULSE under various degradation operators. In particular, these are downscaling followed by Gaussian noise (std=$25,50$), motion blur in random directions with length $100$ followed by downscaling, and downscaling followed by salt-and-pepper noise with a density of $0.05$.}
    \label{fig:NOISE}
\end{figure}

\section{Robustness}
The main aim of our algorithm is to perform perceptually realistic super-resolution with a known downscaling operator. However, we find that even for a variety of \textit{unknown} downscaling operators, we can apply our method using bicubic downscaling as a stand-in for more substantial degradations applied--see Figure \ref{fig:NOISE}. In this case, we provide only the degraded low-resolution image as input. We find that the output downscales approximately to the \textit{true}, non-noisy LR image (that is, the bicubically downscaled HR) rather than to the degraded LR given as input. This is desired behavior, as we would not want to create an image that matches the additional degradations. PULSE thus implicitly denoises images. This is due to the fact that we restrict the outputs to only realistic faces, which in turn can only downscale to reasonable LR faces. Traditional supervised networks, on the other hand, are sensitive to added noise and changes in the domain and must therefore be explicitly trained with the noisy inputs (e.g., \cite{DNSR}). 

\section{Bias} \label{bias}

\begin{table*}[ht]
\centering
\begin{tabular}{|c|c|c|c|c|c|c|}
    \hline
    \multicolumn{7}{|c|}{Race}                                                                 \\ \hline
    Black  & East Asian & Indian & Latino/Hispanic & Middle Eastern & Southeast Asian & White  \\ \hline
    79.2\% & 87.0\%     & 87.4\% & 90.2\%          & 87.0\%         & 87.4\%          & 83.4\% \\ \hline
\end{tabular}
\begin{tabular}{|c|c|}
    \hline
    \multicolumn{2}{|c|}{Gender} \\ \hline
    Female        & Male         \\ \hline
    91.4\%        & 88.6\%       \\ \hline
\end{tabular}
\label{tab:bias}\newline
\caption{Success rates (frequency with which PULSE finds an image in the outputs of the generator that downscales correctly) of PULSE with StyleGAN-FFHQ across various groups, evaluated on FairFace. See ``Failure to converge'' in Section \ref{bias} for full explanation of this analysis and its limitations.}
\end{table*}

While we initially chose to demonstrate PULSE using StyleGAN (trained on FFHQ) as the generative model for its impressive image quality, we noticed some bias when evaluated on natural images of faces outside of our test set. In particular, we believe that PULSE may illuminate some biases inherent in StyleGAN. We document this in a more structured way with a model card in Figure \ref{fig:modelcard}, where we also examine success/failure rates of PULSE across different subpopulations. We propose a few possible sources for this bias:
\newline

\noindent\textit{Bias inherited from latent space constraint:} If StyleGAN pathologically placed people of color in areas of lower density in the latent space, bias would be introduced by PULSE's constraint on the latent space which is necessary to consistently generate high resolution images. To evaluate this, we ran PULSE with different radii for the hypersphere PULSE searches on, corresponding to different samples. This did not seem to have an effect. 
\newline

\noindent\textit{Failure to converge:} In the initial code we released on GitHub, PULSE failed to return ``no image found'' when at the end of optimization it still did not find an image that downscaled correctly (within $\epsilon$). The concern could therefore be that it is harder to find images in the outputs of StyleGAN that downscale to people of color than to white people. To test this, we found a new dataset with better representation to evaluate success/failure rates on, ``FairFace: Face Attribute Dataset for Balanced Race, Gender, and Age'' \cite{fairface}. This dataset was labeled by third-party annotators on these fields. We sample 100 examples per subgroup and calculate the success/failure rate across groups with $\times64$ downscaling after running PULSE 5 times per image. The results of this experiment can be found in Table \ref{tab:bias}. There is some variation in these percentages, but it does not seem to be the primary cause of what we observed. Note that this metric is lacking in that it only reports whether an image was found - which does not reflect the diversity of images found over many runs on many images, an important measure that was difficult to quantify.
\newline

\noindent\textit{Bias inherited from optimization:} This would imply that the constrained latent space contains a wide range of images of people of color but that PULSE's optimization procedure does not find them. However, if this is the case then we should be able to find such images with enough random initializations in the constrained latent space. We ran this experiment and this also did not seem to have an effect. \newline

\noindent\textit{Bias inherited from StyleGAN:} Some have noted that it seems more diverse images can be found in an augmented latent space of StyleGAN per \cite{styleganembedding}. However, this is not close to the set of images StyleGAN itself generates when trained on faces: for example, in the same paper, the authors display images of unrelated domains (such as cats) being embedded successfully as well. In our work, PULSE is constrained to images StyleGAN considers realistic face images (actually, a slight expansion of this set; see below and Appendix for an explanation of this).
    
More technically: in StyleGAN, they sample a latent vector $z$, which is fed through the mapping network to become a vector $w$, which is duplicated 18 times to be fed through the synthesis network. In \cite{styleganembedding}, they instead find 18 different vectors to feed through the synthesis network that correspond to any image of interest (whether faces or otherwise). In addition, while each of these latent vectors would have norm $\approx \sqrt{512}$ when sampled, this augmented latent space allows them to vary freely, potentially finding points in the latent space very far from what would have been seen in training. Using this augmented latent space therefore removes any guarantee that the latent recovered corresponds to a realistic image of a face. 
    
In our work, instead of duplicating the vector 18 times as StyleGAN does, we relax this constraint and encourage these 18 vectors to be approximately equal to each other so as to still generate realistic outputs (see Appendix). This relaxation means the set of images PULSE can generate should be broader than the set of images StyleGAN could produce naturally. We found that loosening any of StyleGAN's constraints on the latent space further generally led to unrealistic faces or images that were not faces.

Overall, it seems that sampling from StyleGAN yields white faces much more frequently than faces of people of color, indicating more of the prior density may be dedicated to white faces. Recent work by Salminen et al. \cite{Salminen_Jung_Chowdhury_Jansen_2020b} describes the implicit biases of StyleGAN in more detail, which seem to confirm these observations. In particular, we note their analysis of the demographic bias of the outputs of the model:
\begin{quote}
“Results indicate a racial bias among the generated pictures, with close to three-[fourths] (72.6\%) of the pictures representing White people. Asian (13.8\%) and Black (10.1\%) are considerably less frequent, while Indians represent only a minor fraction of the pictures (3.4\%).”
\end{quote}
This bias extends to any downstream application of StyleGAN, including the implementation of PULSE using StyleGAN.
%
%We have since contacted the creator of FFHQ and StyleGAN, NVIDIA, to raise this issue. This also highlights the importance of bias evaluation of generative models. Furthermore, it makes the importance of tools like VIBE \cite{wang2020vibe} for automated analysis of dataset bias especially clear as we use bigger and bigger datasets that are infeasible to parse by hand.

%As this work was on image-to-image models of faces, it provides a particularly clear example of biased results from machine learning models; we hope this serves as a reminder that we must be even more careful when working with data that doesn't admit visual inspection so readily (for instance, models that make credit, medical decisions and more), especially for engineers implementing these in the real world.

\section{Discussion and Future Work}
Through these experiments, we find that PULSE produces perceptually superior images that also downscale correctly. PULSE accomplishes this at resolutions previously unseen in the literature. All of this is done with unsupervised methods, removing the need for training on paired datasets of LR-HR images. The visual quality of our images as well as MOS and NIQE scores demonstrate that our proposed formulation of the super-resolution problem corresponds with human intuition. Starting with a pre-trained GAN, our method operates only at test time, generating each image in about 5 seconds on a single GPU. However, we also note significant limitations when evaluated on natural images past the standard benchmark.

One reasonable concern when searching the output space of GANs for images that downscale properly is that while GANs generate sharp images, they need not cover the whole distribution as, e.g., flow-based models must. In our experiments using CelebA and StyleGAN, we did not observe any manifestations of this, which may be attributable to bias – see Section \ref{bias} (The ``mode collapse'' behavior of GANs may exacerbate dataset bias and contribute to the results described in Section \ref{bias} and the model card, Figure \ref{fig:modelcard}.) Advances in generative modeling will allow for generative models with better coverage of larger distributions, which can be directly used with PULSE without modification.

Another potential concern that may arise when considering this unsupervised approach is the case of an unknown downscaling function. In this work, we focused on the most prominent SR use case: on bicubically downscaled images. In fact, in many use cases, the downscaling function is either known analytically (e.g., bicubic) or is a (known) function of hardware. However, methods have shown that the degradations can be estimated in entirely unsupervised fashions for arbitrary LR images (that is, not necessarily those which have been downscaled bicubically)~\cite{bulat2018learn,zhao2018unsupervised}. Through such methods, we can retain the algorithm's lack of supervision; integrating these is an interesting topic for future work.

\section{Conclusions}

We have established a novel methodology for image super-resolution as well as a new problem formulation. This opens up a new avenue for super-resolution methods along different tracks than traditional, supervised work with CNNs. The approach is not limited to a particular degradation operator seen during training, and it always maintains high perceptual quality. 
%Furthermore, our results retain high perceptual quality while generating 64 times more pixels per image compared to previous work.

\vspace*{5pt}
\noindent\textbf{Acknowledgments:}
Funding was provided by the Lord Foundation of North Carolina and the Duke Department of Computer Science. Thank you to the Google Cloud Platform research credits program.
%for enabling us to access Google Cloud Platform computational resources.

{
\bibliographystyle{ieee_fullname}
\bibliography{ref}

\begin{thebibliography}{10}\itemsep=-1pt

\bibitem{abadi2016tensorflow}
Mart{\'\i}n Abadi, Paul Barham, Jianmin Chen, Zhifeng Chen, Andy Davis, Jeffrey
  Dean, Matthieu Devin, Sanjay Ghemawat, Geoffrey Irving, Michael Isard, et~al.
\newblock Tensorflow: A system for large-scale machine learning.
\newblock In {\em 12th $\{$USENIX$\}$ symposium on operating systems design and
  implementation ($\{$OSDI$\}$ 16)}, pages 265--283, 2016.

\bibitem{styleganembedding}
Rameen Abdal, Yipeng Qin, and Peter Wonka.
\newblock {Image2StyleGAN}: How to embed images into the {StyleGAN} latent
  space?
\newblock In {\em Proceedings of the International Conference on Computer
  Vision {(ICCV)}}, 2019.

\bibitem{baker2000limits}
Simon Baker and Takeo Kanade.
\newblock Limits on super-resolution and how to break them.
\newblock In {\em Proceedings IEEE Conference on Computer Vision and Pattern
  Recognition (CVPR)}, volume~2, pages 372--379. IEEE, 2000.

\bibitem{DNSR}
Yijie Bei, Alexandru Damian, Shijia Hu, Sachit Menon, Nikhil Ravi, and Cynthia
  Rudin.
\newblock New techniques for preserving global structure and denoising with low
  information loss in single-image super-resolution.
\newblock In {\em 2018 {IEEE} Conference on Computer Vision and Pattern
  Recognition Workshops, {CVPR} Workshops 2018, Salt Lake City, UT, USA, June
  18-22, 2018}, pages 874--881. {IEEE} Computer Society, 2018.

\bibitem{Blau_2018_ECCV_Workshops}
Yochai Blau, Roey Mechrez, Radu Timofte, Tomer Michaeli, and Lihi Zelnik-Manor.
\newblock The 2018 {PIRM} challenge on perceptual image super-resolution.
\newblock In {\em The European Conference on Computer Vision (ECCV) Workshops},
  September 2018.

\bibitem{bora2017compressed}
Ashish Bora, Ajil Jalal, Eric Price, and Alexandros~G. Dimakis.
\newblock Compressed sensing using generative models.
\newblock In {\em Proceedings of the 34th International Conference on Machine
  Learning (ICML)}, 2017.

\bibitem{bulat2018learn}
Adrian Bulat, Jing Yang, and Georgios Tzimiropoulos.
\newblock To learn image super-resolution, use a gan to learn how to do image
  degradation first.
\newblock In {\em Proceedings of the European Conference on Computer Vision
  (ECCV)}, pages 185--200, 2018.

\bibitem{chen2018fsrnet}
Yu Chen, Ying Tai, Xiaoming Liu, Chunhua Shen, and Jian Yang.
\newblock Fsrnet: End-to-end learning face super-resolution with facial priors.
\newblock In {\em Proceedings of the IEEE Conference on Computer Vision and
  Pattern Recognition {(CVPR)}}, pages 2492--2501, 2018.

\bibitem{10.1007/978-3-319-10593-2_13}
Chao Dong, Chen~Change Loy, Kaiming He, and Xiaoou Tang.
\newblock Learning a deep convolutional network for image super-resolution.
\newblock In {\em Proceedings of the European Conference on Computer Vision
  {(ECCV)}}, pages 184--199, 2014.

\bibitem{goodfellow2014generative}
Ian Goodfellow, Jean Pouget-Abadie, Mehdi Mirza, Bing Xu, David Warde-Farley,
  Sherjil Ozair, Aaron Courville, and Yoshua Bengio.
\newblock Generative adversarial nets.
\newblock In {\em Advances in Neural Information Processing Systems}, pages
  2672--2680, 2014.

\bibitem{jeong2015multi}
Seokhwa Jeong, Inhye Yoon, and Joonki Paik.
\newblock Multi-frame example-based super-resolution using locally directional
  self-similarity.
\newblock {\em IEEE Transactions on Consumer Electronics}, 61(3):353--358,
  2015.

\bibitem{karras2017progressive}
Tero Karras, Timo Aila, Samuli Laine, and Jaakko Lehtinen.
\newblock Progressive growing of {GAN}s for improved quality, stability, and
  variation.
\newblock {\em CoRR}, abs/1710.10196, 2018.
\newblock Appeared at the 6th Annual International Conference on Learning
  Representations.

\bibitem{karras2019style}
Tero Karras, Samuli Laine, and Timo Aila.
\newblock A style-based generator architecture for generative adversarial
  networks.
\newblock In {\em Proceedings of the IEEE Conference on Computer Vision and
  Pattern Recognition {(CVPR)}}, pages 4401--4410, 2019.

\bibitem{kim2019progressive}
Deokyun Kim, Minseon Kim, Gihyun Kwon, and Dae-Shik Kim.
\newblock Progressive face super-resolution via attention to facial landmark.
\newblock In {\em Proceedings of the 30th British Machine Vision Conference
  {(BMVC)}}, 2019.

\bibitem{VDSR}
Jiwon Kim, Jung~Kwon Lee, and Kyoung~Mu Lee.
\newblock Accurate image super-resolution using very deep convolutional
  networks.
\newblock In {\em Proceedings of IEEE Conference on Computer Vision and Pattern
  Recognition ({CVPR})}, 2016.

\bibitem{SRGAN}
Christian Ledig, Lucas Theis, Ferenc Husz{\'a}r, Jose Caballero, Andrew
  Cunningham, Alejandro Acosta, Andrew Aitken, Alykhan Tejani, Johannes Totz,
  Zehan Wang, et~al.
\newblock Photo-realistic single image super-resolution using a generative
  adversarial network.
\newblock In {\em Proceedings of the IEEE Conference on Computer Vision and
  Pattern Recognition {(CVPR)}}, pages 4681--4690, 2017.

\bibitem{modelcards}
Margaret Mitchell, Simone Wu, Andrew Zaldivar, Parker Barnes, Lucy Vasserman,
  Ben Hutchinson, Elena Spitzer, Inioluwa~Deborah Raji, and Timnit Gebru.
\newblock Model cards for model reporting.
\newblock In {\em Proceedings of the conference on fairness, accountability,
  and transparency}, pages 220--229, 2019.

\bibitem{NIQE}
A. {Mittal}, R. {Soundararajan}, and A.~C. {Bovik}.
\newblock Making a “completely blind” image quality analyzer.
\newblock {\em IEEE Signal Processing Letters}, 20(3):209--212, March 2013.

\bibitem{paszke2019pytorch}
Adam Paszke, Sam Gross, Francisco Massa, Adam Lerer, James Bradbury, Gregory
  Chanan, Trevor Killeen, Zeming Lin, Natalia Gimelshein, Luca Antiga, et~al.
\newblock Pytorch: An imperative style, high-performance deep learning library.
\newblock In {\em Advances in neural information processing systems}, pages
  8026--8037, 2019.

\bibitem{unet}
Olaf Ronneberger, Philipp Fischer, and Thomas Brox.
\newblock U-net: Convolutional networks for biomedical image segmentation.
\newblock In {\em Proceedings of the International Conference on Medical Image
  Computing and Computer-Assisted Intervention}, pages 234--241. Springer,
  2015.

\bibitem{Salminen_Jung_Chowdhury_Jansen_2020b}
Joni Salminen, Soon-gyo Jung, Shammur Chowdhury, and Bernard~J. Jansen.
\newblock Analyzing demographic bias in artificially generated facial pictures.
\newblock In {\em Extended Abstracts of the 2020 CHI Conference on Human
  Factors in Computing Systems}, page 1–8. ACM, Apr 2020.

\bibitem{Shi_2016_CVPR}
Wenzhe Shi, Jose Caballero, Ferenc Huszar, Johannes Totz, Andrew~P. Aitken, Rob
  Bishop, Daniel Rueckert, and Zehan Wang.
\newblock Real-time single image and video super-resolution using an efficient
  sub-pixel convolutional neural network.
\newblock In {\em Proceedings of the IEEE Conference on Computer Vision and
  Pattern Recognition (CVPR)}, June 2016.

\bibitem{actrec}
Karen Simonyan and Andrew Zisserman.
\newblock Two-stream convolutional networks for action recognition in videos.
\newblock In {\em Advances in Neural Information Processing Systems}, pages
  568--576, 2014.

\bibitem{singh2016super}
Amanjot Singh and Jagroop~Singh Sidhu.
\newblock Super resolution applications in modern digital image processing.
\newblock {\em International Journal of Computer Applications},
  150(2):0975--8887, 2016.

\bibitem{deep_prior}
Dmitry Ulyanov, Andrea Vedaldi, and Victor Lempitsky.
\newblock Deep image prior.
\newblock In {\em Proceedings of the IEEE Conference on Computer Vision and
  Pattern Recognition (CVPR)}, June 2018.

\bibitem{vershynin_2018}
Roman Vershynin.
\newblock {\em Random Vectors in High Dimensions}, page 38–69.
\newblock Cambridge Series in Statistical and Probabilistic Mathematics.
  Cambridge University Press, 2018.

\bibitem{PyTorchStyleGAN}
Thomas Viehmann and Lernapparat.
\newblock Pytorch implementation of the stylegan generator.
\newblock \url{https://github.com/lernapparat/lernapparat/}, 2019.

\bibitem{srsurvey}
Nannan Wang, Dacheng Tao, Xinbo Gao, Xuelong Li, and Jie Li.
\newblock A comprehensive survey to face hallucination.
\newblock {\em International Journal of Computer Vision}, 106(1):9--30, 2014.

\bibitem{esrgan}
Xintao Wang, Ke Yu, Shixiang Wu, Jinjin Gu, Yihao Liu, Chao Dong, Yu Qiao, and
  Chen Change~Loy.
\newblock {ESRGAN}: Enhanced super-resolution generative adversarial networks.
\newblock In {\em Proceedings of the European Conference on Computer Vision
  {(ECCV)} Workshops}, September 2018.

\bibitem{fairface}
D. {Xu}, S. {Yuan}, L. {Zhang}, and X. {Wu}.
\newblock Fairgan+: Achieving fair data generation and classification through
  generative adversarial nets.
\newblock In {\em 2019 IEEE International Conference on Big Data (Big Data)},
  pages 1401--1406, 2019.

\bibitem{zhao2018unsupervised}
Tianyu Zhao, Changqing Zhang, Wenqi Ren, Dongwei Ren, and Qinghua Hu.
\newblock Unsupervised degradation learning for single image super-resolution.
\newblock {\em arXiv preprint arXiv:1812.04240}, 2018.

\end{thebibliography}


\begin{thebibliography}{1}\itemsep=-1pt

\bibitem{styleganembedding}
Rameen Abdal, Yipeng Qin, and Peter Wonka.
\newblock {Image2StyleGAN}: How to embed images into the {StyleGAN} latent
  space?
\newblock In {\em Proceedings of the International Conference on Computer
  Vision {(ICCV)}}, 2019.

\end{thebibliography}
}

\pagebreak

\begin{figure*}[ht]  \label{fig:modelcard} 
\begin{tcolorbox}[
    sharp corners,
    colback=white,
    colframe=black
    ]
\begin{center}
\Large\textbf{Model Card - PULSE with StyleGAN FFHQ Generative Model Backbone} \\
\end{center}
\small
\textbf{Model Details}
\begin{itemize}
    \item PULSE developed by researchers at Duke University, 2020, v1.
    \begin{itemize}
        \item Latent Space Exploration Technique.
        \item PULSE does no training, but is evaluated by downscaling loss (equation \ref{eq:dsloss}) for fidelity to input low-resolution image.
        \item Requires pre-trained generator to parameterize natural image manifold.
    \end{itemize}
    \item StyleGAN developed by researchers at NVIDIA, 2018, v1.
    \begin{itemize}
        \item Generative Adversarial Network.
        \item StyleGAN trained with adversarial loss (WGAN-GP).
    \end{itemize}
\end{itemize}
\textbf{Intended Use}
\begin{itemize}
    \item PULSE was intended as a proof of concept for one-to-many super-resolution (generating multiple high resolution outputs from a single image) using latent space exploration.
    \item Intended use of implementation using StyleGAN-FFHQ (faces) is purely as an art project - seeing fun pictures of imaginary people that downscale approximately to a low-resolution image.
    \item Not suitable for facial recognition/identification. PULSE makes imaginary faces of people who do not exist, which should not be confused for real people. It will not help identify or reconstruct the original image.
    \item Demonstrates that face recognition is not possible from low resolution or blurry images because PULSE can produce visually distinct high resolution images that all downscale correctly.
\end{itemize}
\textbf{Factors}
\begin{itemize}
    \item Similarly to \cite{modelcards}: ``based on known problems with computer vision face technology, potential relevant factors include groups for gender, age, race, and Fitzpatrick skin type.'' Additional factors include lighting, background content, hairstyle, pose, camera focal length, and accessories.
    %\item Similarly to \cite{modelcards}: “based on known problems with computer vision face technology, potential relevant factors include groups for gender, age, race, and Fitzpatrick skin type; hardware factors of camera type and lens type; and environmental factors of lighting and humidity.” These potentially relevant factors extend here as well. Further relevant factors important to this work include background content, hairstyle, pose, and accessories (such as glasses, hats, earrings, and scarves).
    % \item We follow the CelebA model card example in \cite{modelcards}: “Evaluation factors are gender and age group, as annotated in the publicly available dataset CelebA [CITE CELEBA]. Further possible factors not currently available in a public [high definition facial image dataset]. Gender and age determined by third-party annotators based on visual presentation, following a set of examples of male/female gender and young/old age. Further details available in [CITE CELEBA].”
\end{itemize}
\textbf{Metrics}
\begin{itemize}
    \item Evaluation metrics include the success/failure rate of PULSE using StyleGAN (when it does not find an image that downscales appropriately). Note that this metric is lacking in that it only reports whether an image was found - which does not reflect the diversity of images found over many runs on many images, an important measure that was difficult to quantify. In original evaluation experiments, the failure rate was zero.
    
    \item To better evaluate the way that this metric varies across groups, we found a new dataset with better representation to evaluate success/failure rates on, ``FairFace: Face Attribute Dataset for Balanced Race, Gender, and Age'' \cite{fairface}. This dataset was labeled by third-party annotators on these fields. We sample 100 examples per subgroup and calculate the success/failure rate across groups with $\times64$ downscaling after running PULSE 5 times per image. We note that this is a small sample size per subgroup. The results of this analysis can be found in Table \ref{tab:bias}.

    %\item Evaluation metrics for fairness in generative models were difficult to find. However, given the reliance of PULSE on a generative model backbone, it is crucial to have some understanding of the biases of the model it uses, since PULSE propagates those biases. Thus, we refer to \cite{Salminen_Jung_Chowdhury_Jansen_2020b} to report some metrics on StyleGAN-FFHQ.
\end{itemize}
\vspace{-12pt}
\begin{multicols}{2}
\textbf{Training Data}
\begin{itemize}
    \item PULSE is never trained itself, it leverages a pretrained generator.
    \item StyleGAN is trained on FFHQ \cite{karras2019style}. 
\end{itemize}
\columnbreak
\textbf{Evaluation Data}
\begin{itemize}
    \item CelebA HQ, test data split, chosen as a basic proof of concept. 
    \item MOS Score evaluated via ratings from third party annotators (Amazon MTurk)
\end{itemize}
\end{multicols}
\vspace{-12pt}
\textbf{Ethical Considerations}
\begin{itemize}
    \item Evaluation data - CelebA HQ: Faces based on public figures (celebrities). Only image data is used (no additional annotations). However, we point out that this dataset has been noted to have a severe imbalance of white faces compared to faces of people of color (almost 90\% white) \cite{fairface}. This leads to evaluation bias. As this has been the accepted benchmark in face super-resolution, issues of bias in this field may go unnoticed.
\end{itemize}
\textbf{Caveats and Recommendations}
\begin{itemize}
    \item FairFace appears to be a better dataset to use than CelebA HQ for evaluation purposes.
    \item Due to lack of available compute, we could not at this time analyze intersectional identities and the associated biases.
    \item For an in depth discussion of the biases of StyleGAN, see \cite{Salminen_Jung_Chowdhury_Jansen_2020b}.
    \item Finally, again similarly to \cite{modelcards}:
    \begin{enumerate}
    \item “Does not capture race or skin type, which has been reported as a source of disproportionate errors.
    \item Given gender classes are binary (male/not male), which we include as male/female. Further work needed to evaluate across a
    spectrum of genders.
    \item An ideal evaluation dataset would additionally include annotations for Fitzpatrick skin type, camera details, and environment (lighting/humidity) details.”
\end{enumerate}
\end{itemize}
\end{tcolorbox}
\end{figure*}

\end{document}

%%%%%%%%% TITLE
\title{PULSE: Supplementary Material}

\author{Sachit Menon*, Alexandru Damian*, Nikhil Ravi, Shijia Hu, Cynthia Rudin\\
Duke University\\
Durham, NC\\
{\tt\small \{sachit.menon,alexandru.damian,nikhil.ravi,shijia.hu,cynthia.rudin\}@duke.edu}
}

\maketitle

\section{Appendix A: Additional Figures}

\begin{figure*}[]
    \centering
    \includegraphics[width=0.95\textwidth]{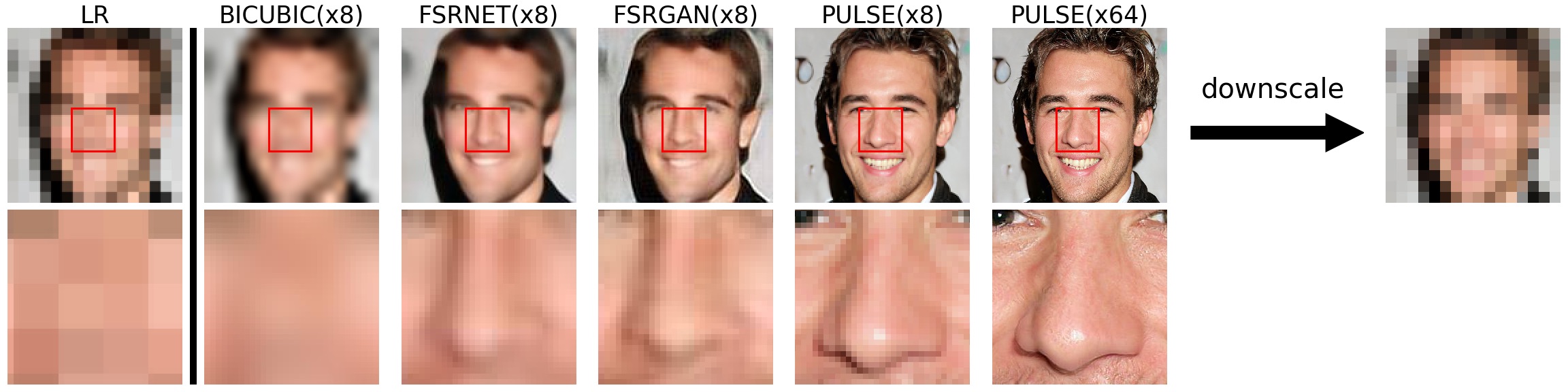}
    \includegraphics[width=0.95\textwidth]{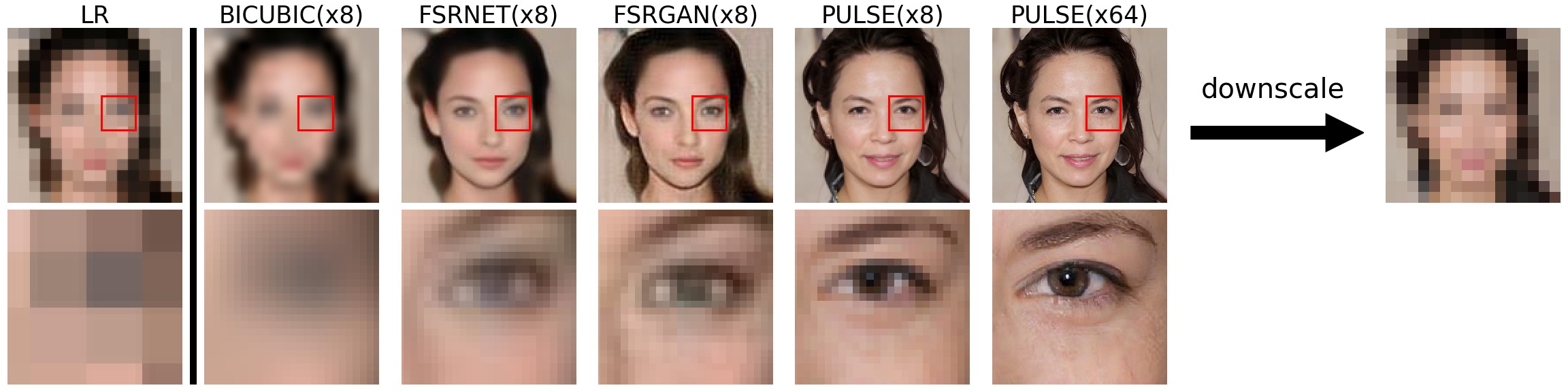}
    \caption{Further comparison of PULSE with bicubic upscaling, FSRNet, and FSRGAN.}
    \label{fig:furtheralgcomparison}
\end{figure*}

\begin{figure*}[]
    \centering
    \includegraphics[width=0.49\textwidth]{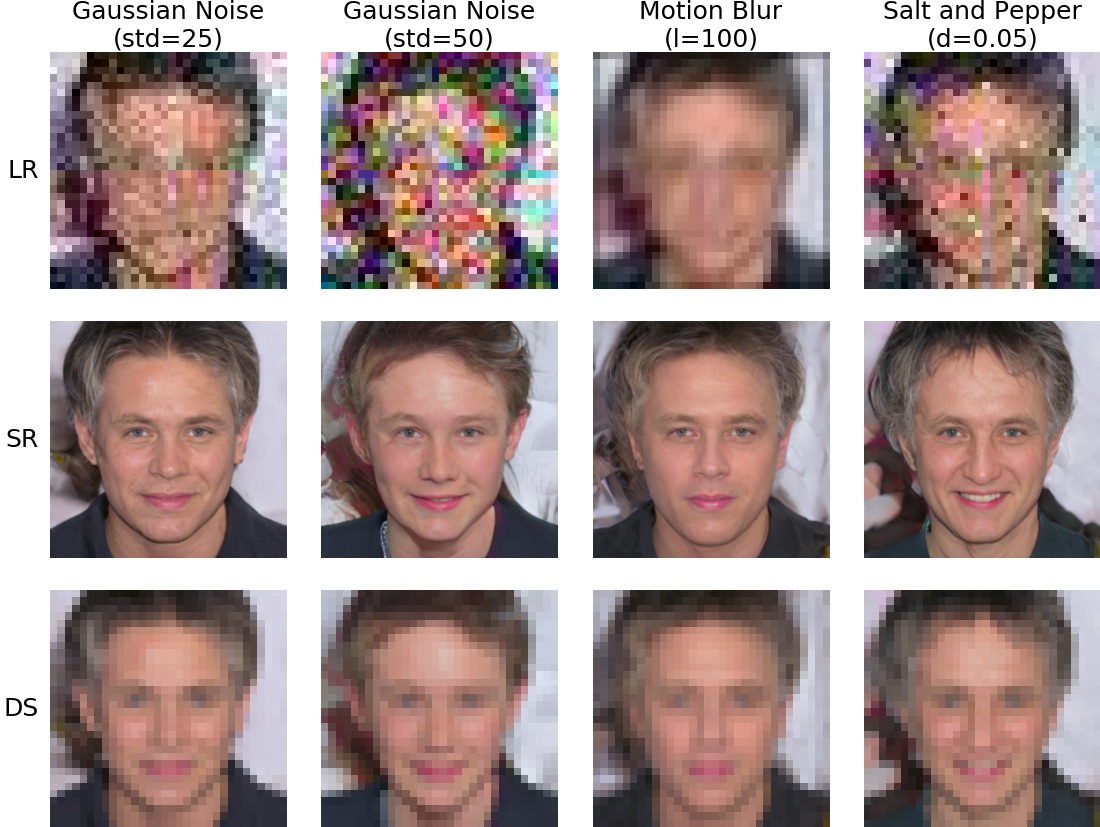}
    \includegraphics[width=0.49\textwidth]{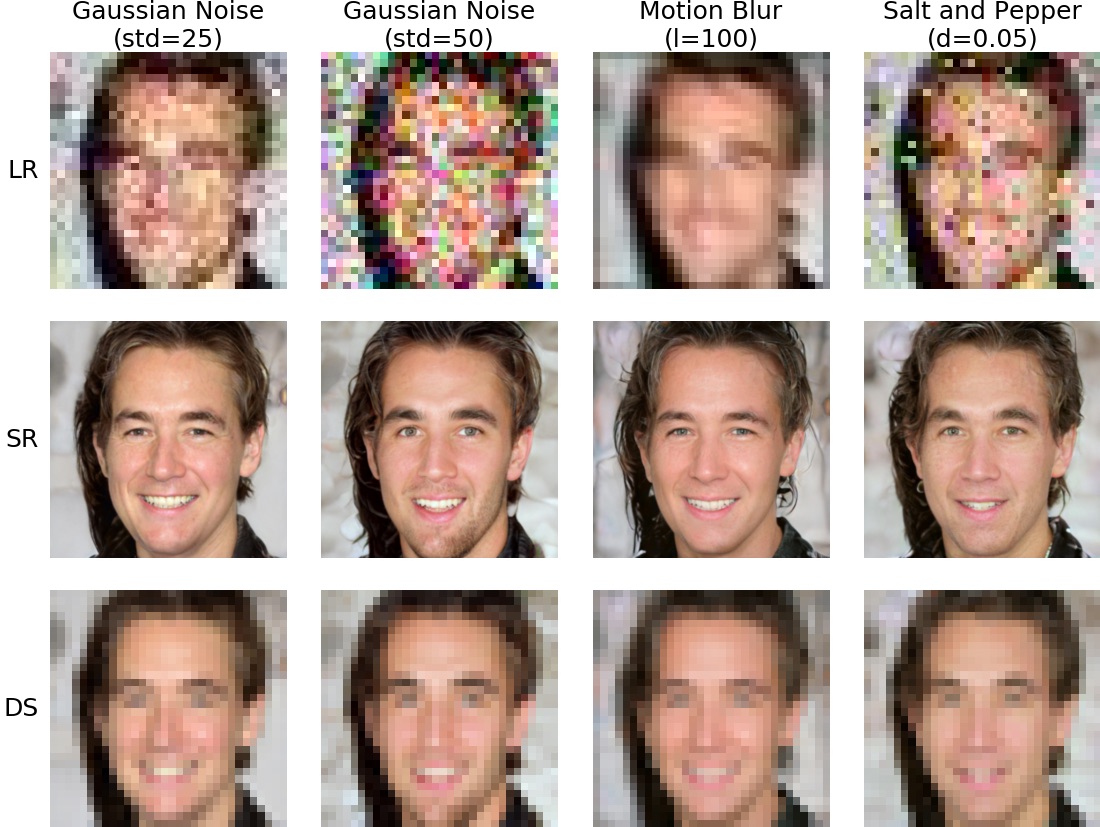}
    \includegraphics[width=0.49\textwidth]{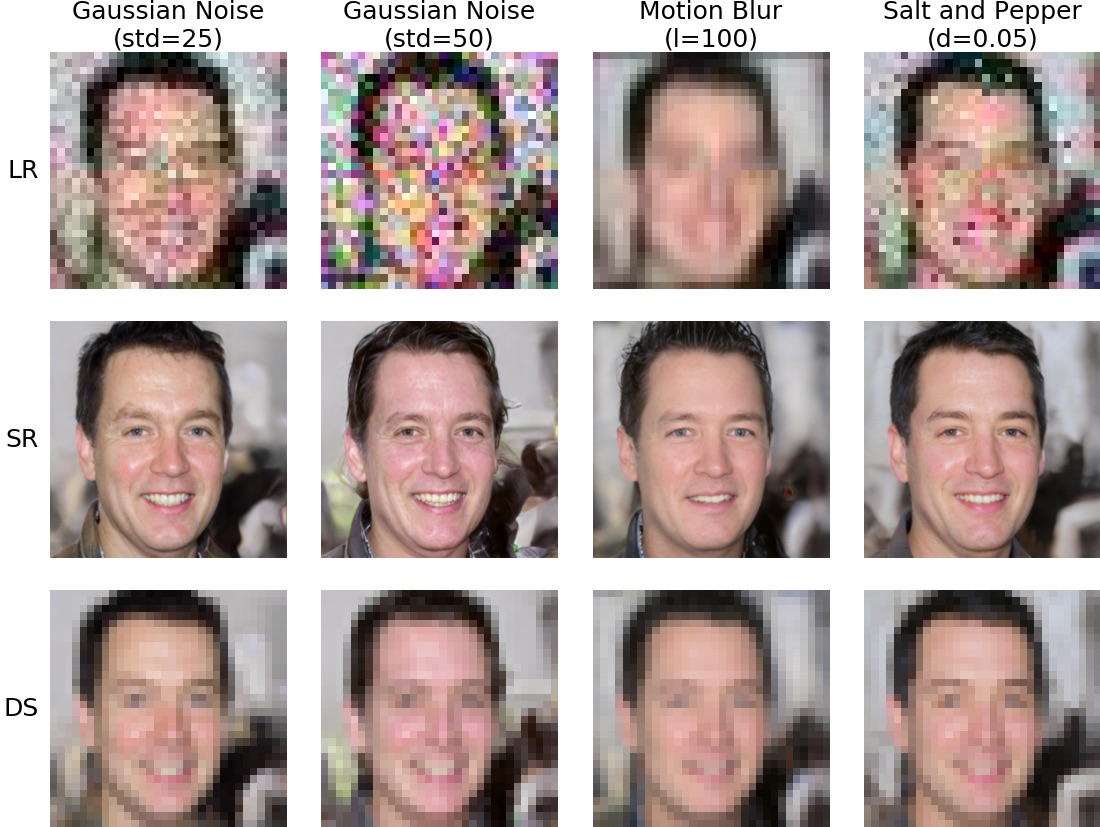}
    \includegraphics[width=0.49\textwidth]{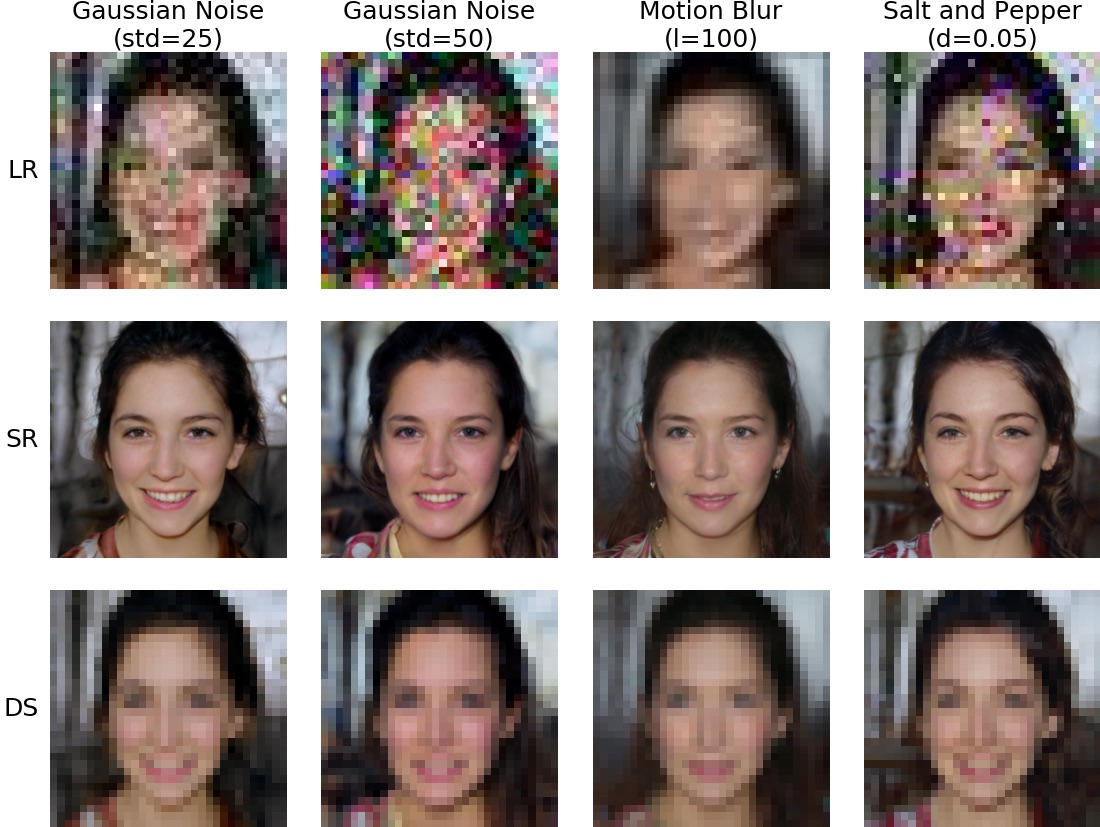}
    \caption{(x32) Additional robustness results for PULSE under additional degradation operators (these are downscaling followed by Gaussian noise (std=$25,50$), motion blur in random directions with length $100$ followed by downscaling, and downscaling followed by salt-and-pepper noise with a density of $0.05$.)}
    \label{fig:furthernoisecomparison}
\end{figure*}

\begin{figure*}[ht!]
    \centering
    \includegraphics[width=0.8\textwidth]{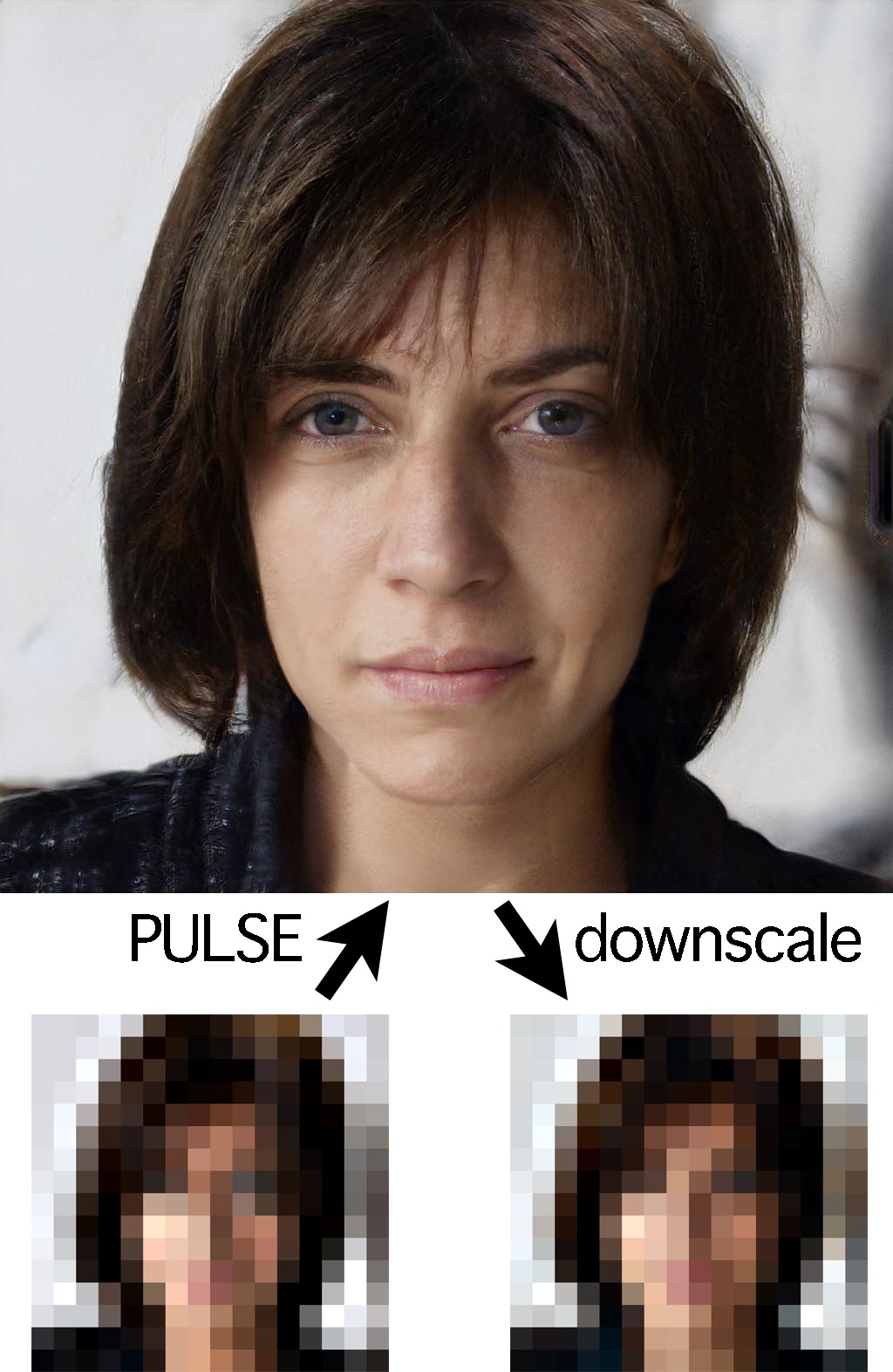}
    \caption{(64x) Sample 1}
    \label{fig:samples}
\end{figure*}

\begin{figure*}[ht!]
    \centering
    \includegraphics[width=0.8\textwidth]{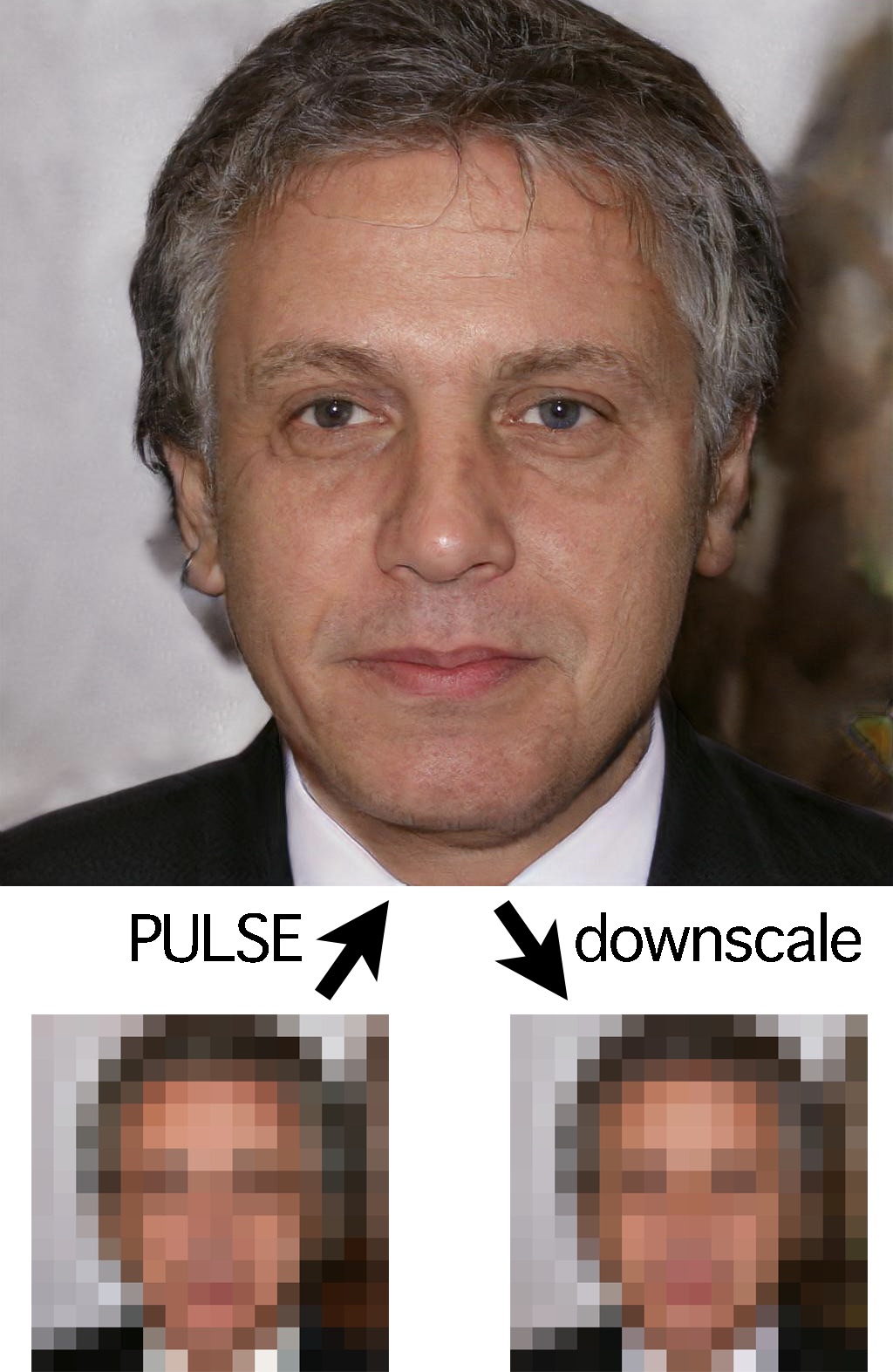}
    \caption{(64x) Sample 2}
    \label{fig:samples2}
\end{figure*}

\begin{figure*}[ht!]
    \centering
    \includegraphics[width=0.8\textwidth]{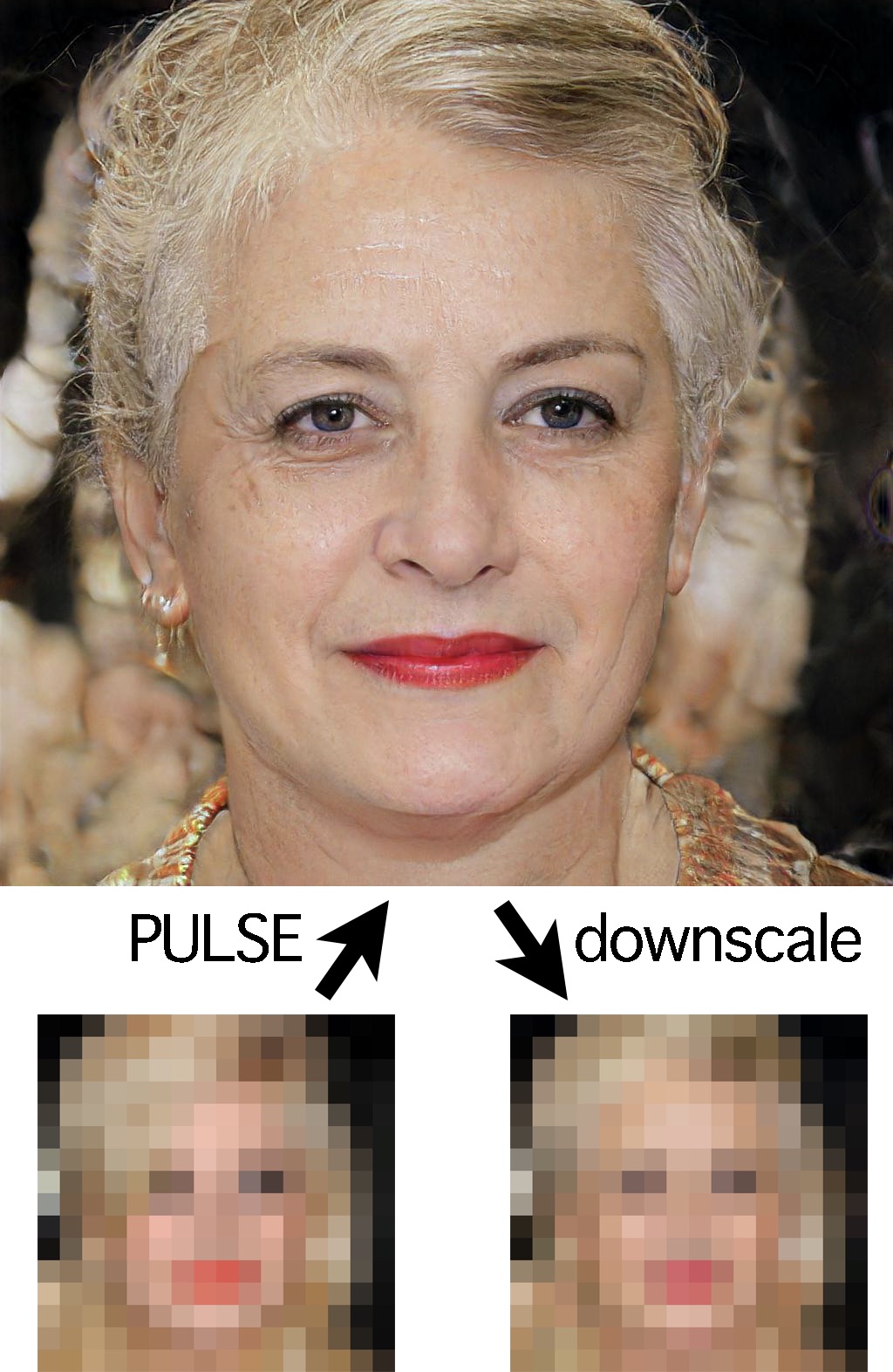}
    \caption{(64x) Sample 3}
    \label{fig:samples3}
\end{figure*}

\begin{figure*}[ht!]
    \centering
    \includegraphics[width=0.8\textwidth]{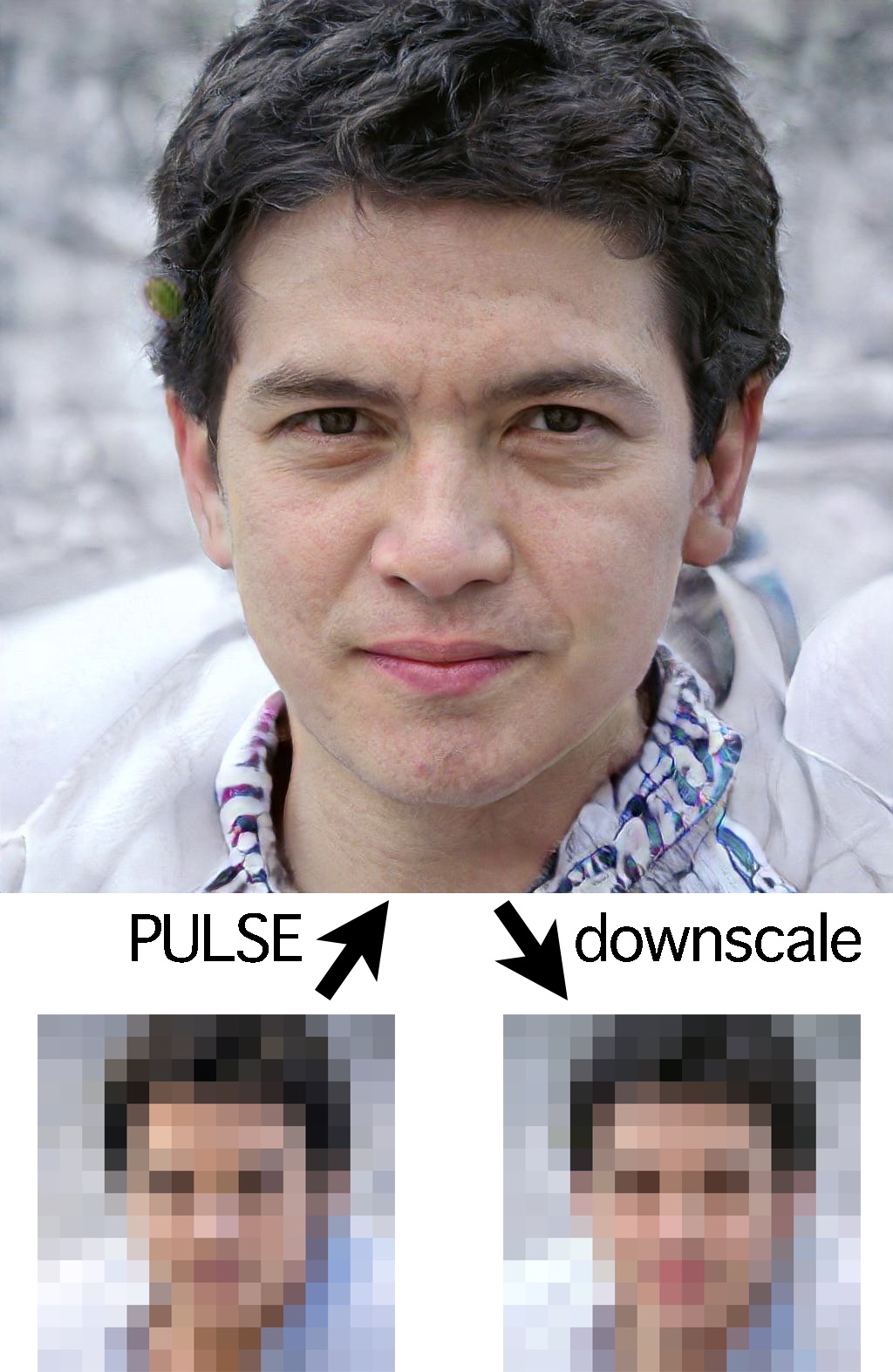}
    \caption{(64x) Sample 4}
    \label{fig:samples4}
\end{figure*}

Here we provide further samples of the output of our super-resolution method for illustration in Figure \ref{fig:furtheralgcomparison}. These results were obtained with $\times 8$ scale factor from an input of resolution $16 \times 16$. This highlights our method's capacity to illustrate detailed features that we did not have space to show in the main paper, such as noses and a wider variety of eyes. We also present additional examples depicting PULSE's robustness to additional degradation operators in Figure \ref{fig:furthernoisecomparison} and some randomly selected generated samples in Figures \ref{fig:samples}, \ref{fig:samples2}, \ref{fig:samples3}, and \ref{fig:samples4}.

\section{Appendix B: Implementation Details}

\subsection{StyleGAN}

In order to generate experimental results using our method, we had to pick a pretrained generative model to work with. For this we chose StyleGAN due to its state-of-the-art performance on high resolution image generation.

StyleGAN consists of two components: first, a mapping network $M: \mathbb{R}^{512} \rightarrow \mathbb{R}^{512}$, a tiling function $T: \mathbb{R}^{512} \rightarrow \mathbb{R}^{18 \times 512}$, and a synthesis network $S: \mathbb{R}^{18 \times 512} \times \mathcal{N} \rightarrow \mathbb{R}^{1024 \times 1024}$ where $\mathcal{N}$ is collection of Euclidean spaces of varying dimensions representing the domains of each of the noise vectors fed into the synthesis network. To generate images, a vector $z$ is sampled uniformly at random from the surface of the unit sphere in $\mathbb{R}^{512}$. This is transformed into another $512$-dimensional vector by the mapping network, which is replicated $18$ times by the tiling function $T$. The new $18 \times 512$ dimensional vector is input to the synthesis network which uses it to generate a high-resolution, $1024 \times 1024$ pixel image. More precisely, the synthesis network consists of $18$ sequential layers, and the resolution of the generated image is doubled after every other layer; each of these $18$ layers is re-fed into the $512$-dimensional output of the mapping network, hence the tiling function. The synthesis network also takes as input noise sampled from the unit Gaussian, which it uses to stochastically add details to the generated image. Formally, $\eta$ is sampled from the Gaussian prior on $\mathcal{N}$, at which point the output is obtained by computing $S(T(M(z)),\eta)$.

\subsection{Latent Space Embedding}

Experimentally, we observed that optimizing directly on $z \in S^{511} \subset \mathbb{R}^{512}$ yields poor results; this latent space is not expressive enough to map to images that downscale correctly. A logical next step would be to use the expanded $18 \times 512$-dimensional latent space that the synthesis network takes as input, as noted by Abdal, Qin, and Wonka \cite{styleganembedding}. By ignoring the mapping network, $S$ can be applied to any vector in $\mathbb{R}^{18 \times 512}$, rather than only those consisting of a single $512$-dimensional vector repeated $18$ times. This expands the expressive potential of the network; however, by allowing the $18$ $512$-dimensional input vectors to $S$ to vary independently, the synthesis network is no longer constrained to the original domain of StyleGAN.

\subsection{Cross Loss}

For the purposes of super-resolution, such approaches are problematic because they void the guarantee that the algorithm is traversing a good approximation of $\mathcal{M}$, the natural image manifold. The synthesis network was trained with a limited subset of $R^{18 \times 512}$ as input; the further the input it receives is from that subset, the less we know about the output it will produce. The downscaling loss, defined in the main paper, is alone not enough to guide PULSE to a realistic image (only an image that downscales correctly). Thus, we want to make some compromise between the vastly increased expressive power of allowing the input vectors to vary independently and the realism produced by tiling the input to $S$ $18$ times. Instead of optimizing on downscaling loss alone, we need some term in the loss discouraging straying too far in the latent space from the original domain. 

To accomplish this, we introduce another metric, the ``cross loss.'' For a set of vectors $v_1,...,v_k$, we define the cross loss of $v_1,...,v_{k}$ to be
\[
CROSS(v_1,...,v_k) = \sum_{i < j} |v_i-v_j|_2^2
\]

The cross loss imposes a penalty based on the Euclidean distance between every pair of vectors input to $S$. This can be considered a simple form of relaxation on the original constraint that the $18$ input vectors be exactly equal. 

When $v_1,...,v_k$ are sampled from a sphere, it makes more sense to compare geodesic distances along the sphere. This is the approach we used in generating our results. Let $\theta(v,w)$ denote the angle between the vectors $v$ and $w$. We then define the geodesic cross loss to be

\[
GEOCROSS(v_1,...,v_k) = \sum_{i < j} \theta(v_i,v_j)^2
\].

Empirically, by allowing the $18$ input vectors to $S$ to vary while applying the soft constraint of the (geo)cross loss, we can increase the expressive potential of the network without large deviations from the natural image manifold.

\subsection{Approximating the input distribution of $S$}

StyleGAN begins with a uniform distribution on $S^{511} \subset \mathbb{R}^{512}$, which is pushed forward by the mapping network to a transformed probability distribution over $\mathbb{R}^{512}$. Therefore, another requirement to ensure that $S([v_1,...,v_{18}],\eta)$ is a realistic image is that each $v_i$ is sampled from this push-forward distribution. While analyzing this distribution, we found that we could transform this back to a distribution on the unit sphere without the mapping network by simply applying a single linear layer with a leaky-ReLU activation--an entirely invertible transformation. We therefore inverted this function to obtain a sampling procedure for this distribution. First, we generate a latent $w$ from $S^{511}$, and then apply the inverse of our transformation.

\subsection{Noise Inputs}

The second parameter of $S$ controls the stochastic variation that StyleGAN adds to an image. When the noise is set to $0$, StyleGAN generates smoothed-out, detail-free images. The synthesis network takes $18$ noise vectors at varying scales, one at each layer. The earlier noise vectors influence more global features, for example the shape of the face, while the later noise vectors add finer details, such as hair definition. Our first approach was to sample the noise vectors before we began traversing the natural image manifold, keeping them fixed throughout the process. In an attempt to increase the expressive power of the synthesis network, we also tried to perform gradient descent on both the latent input and the noise input to $S$ simultaneously, but this tended to take the noise vectors out of the spherical shell from which they were sampled and produced unrealistic images. Using a standard Gaussian prior forced the noise vectors towards $0$ as mentioned in the main paper. We therefore experimented with two approaches for the noise input:

\begin{enumerate}
    \item \textit{Fixed noise:} Especially when upsampling from $16 \times 16$ to $1024 \times 1024$, StyleGAN was already expressive enough to upsample our images correctly and so we did not need to resort to more complicated techniques.
    
    \item \textit{Partially trainable noise:} In order to slightly increase the expressive power of the network, we optimized on the latent and the first 5-7 noise vectors, allowing us to slightly modify the facial structure of the images we generated to better match the LR images while maintaining image quality. This was the approach we used to generate the images presented in this paper.
\end{enumerate}

\section{Appendix C: Alternative Metrics}

For completeness, we provide the metrics of PSNR and SSIM here. These results were obtained with $\times 8$ scale factor from an input of resolution $16 \times 16$. Note that we explicitly do not aim to optimize on this pixel-wise average distance from the high-resolution image, so these metrics do not have meaningful implications for our work. 

\begin{table}[]
\small{
\renewcommand{\arraystretch}{1.2}
\begin{tabular}{c|c|c|c|c|c|}
\cline{2-6}
                           & Nearest & Bicubic & FSRNet & FSRGAN & PULSE \\ \hline \cline{2-6}
\multicolumn{1}{|c|}{PSNR} & 21.78   & 23.40   & 25.93  & 24.55  & 22.01 \\ \hline
\multicolumn{1}{|c|}{SSIM} & 0.51    & 0.63    & 0.74   & 0.66   & 0.55  \\ \hline
\end{tabular}
}\vspace*{20pt}
\end{table}

\section{Appendix D: Robustness}

Traditional supervised approaches using CNNs are notoriously sensitive to tiny changes in the input domain, as any perturbations are propagated and amplified through the network. This caused some problems when trying to train FSRNET and FSRGAN on FFHQ and then test them on CelebAHQ. However, PULSE never feeds an \LR image through a convolutional network and never applies filters to the LR input images. Instead, the \LR image is only used as a term in the \textit{downscaling loss}. Because the generator is not capable of producing ``noisy'' images, it will seek an \SR image that downscales to the closest point on the \LR natural image manifold to the noisy \LR input. This means that PULSE outputs an \SR image that downscales to the projection of the noisy \LR input onto the \LR natural image manifold, and if the noise is not too strong, this should be close to the ``true'' unperturbed \LR. This may explain why PULSE had no problems when applied to different domains, and why we could demonstrate robustness when the low resolution image was degraded with various types of noise.

{\small
\bibliographystyle{ieee_fullname}
\bibliography{ref}
}